\documentclass{article}
\usepackage{times}
\usepackage[final]{corl_2025}

\usepackage[numbers]{natbib}
\usepackage{multicol}
\usepackage[normalem]{ulem}
\usepackage{xcolor}
\usepackage{url}
\usepackage{graphicx}
\usepackage{float}
\usepackage{tabularx}
\usepackage{wrapfig}
\usepackage{amsmath}
\usepackage{arydshln}
\usepackage{xcolor}
\usepackage{amssymb}
\usepackage{booktabs}
\usepackage{multirow}
\usepackage{enumerate}
\usepackage{afterpage}
\usepackage{placeins}

\newcommand{\name}{\textsc{SocialNav-SUB}}

\title{SocialNav-SUB: Benchmarking VLMs for Scene Understanding in Social Robot Navigation}

\author{
\textbf{Michael J.~Munje\textsuperscript{1}\thanks{Correspondence to: \texttt{michaelmunje@utexas.edu}}, Chen Tang\textsuperscript{1}, Shuijing Liu\textsuperscript{1}, Zichao Hu\textsuperscript{1}, Yifeng Zhu\textsuperscript{1},}\\
\textbf{Jiaxun Cui\textsuperscript{1}, Garrett Warnell\textsuperscript{1,2}, Joydeep Biswas\textsuperscript{1}, Peter Stone\textsuperscript{1,3}}\\
\textsuperscript{1}Department of Computer Science, The University of Texas at Austin\\
\textsuperscript{2}Army Research Laboratory
\textsuperscript{3}Sony AI 
}

\begin{document}
\maketitle

\begin{abstract}
Robot navigation in dynamic, human-centered environments requires socially-compliant decisions grounded in robust scene understanding. Recent Vision-Language Models (VLMs) exhibit promising capabilities such as object recognition, common-sense reasoning, and contextual understanding—capabilities that align with the nuanced requirements of social robot navigation. However, it remains unclear whether VLMs can accurately understand complex social navigation scenes (e.g., inferring the spatial-temporal relations among agents and human intentions), which is essential for safe and socially compliant robot navigation. While some recent works have explored the use of VLMs in social robot navigation, no existing work systematically evaluates their ability to meet these necessary conditions. In this paper, we introduce the \textit{Social Navigation Scene Understanding Benchmark (SocialNav-SUB)}, a Visual Question Answering (VQA) dataset and benchmark designed to evaluate VLMs for scene understanding in real-world social robot navigation scenarios. SocialNav-SUB provides a unified framework for evaluating VLMs against human and rule-based baselines across VQA tasks requiring spatial, spatiotemporal, and social reasoning in social robot navigation. Through experiments with state-of-the-art VLMs, we find that while the best-performing VLM achieves an encouraging probability of agreeing with human answers, it still underperforms simpler rule-based approach and human consensus baselines, indicating critical gaps in social scene understanding of current VLMs. Our benchmark sets the stage for further research on foundation models for social robot navigation, offering a framework to explore how VLMs can be tailored to meet real-world social robot navigation needs. An overview of this paper along with the code and data can be found at \url{https://larg.github.io/socialnav-sub}.
\end{abstract}

\section{Introduction}
\begin{figure}[!htb]
    \centering
    \includegraphics[width=0.9\linewidth]{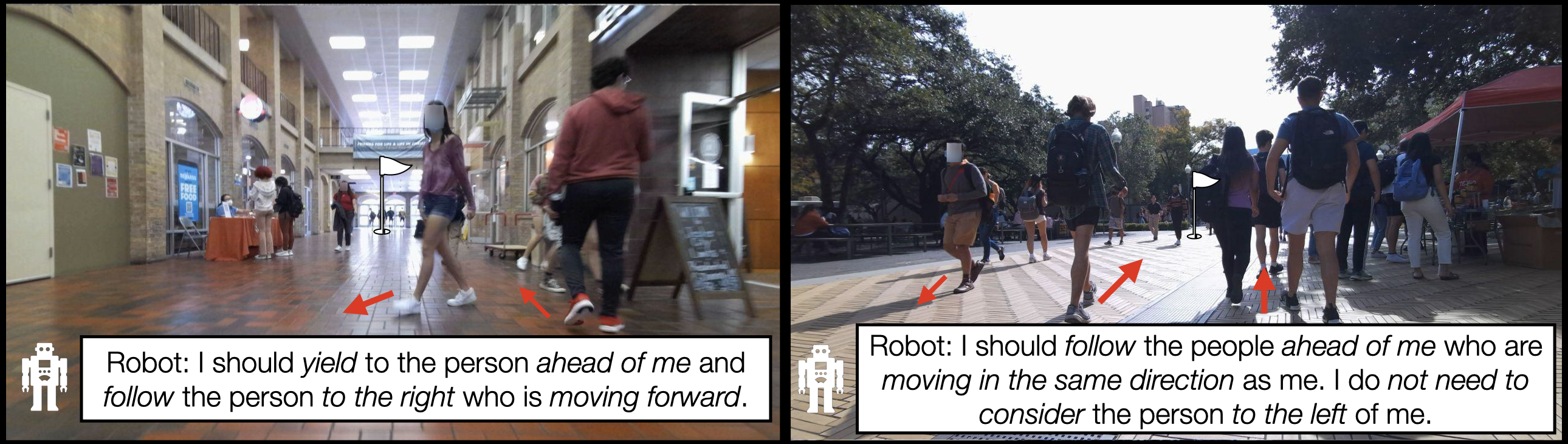}
    \caption{\textbf{Examples of social robot navigation scenarios from SCAND \cite{scand}.} The ability to determine socially compliant navigation actions requires understanding each dynamic scene by spatiotemporal reasoning (e.g. the movements of people in the scene) and social reasoning (inferring the navigation intentions of people in the scene).
    }
    \vspace{-8pt}
    \label{fig:scenario_example}
\end{figure}

\vspace{-5pt}
Social robot navigation, defined as the ability for robots to move effectively and safely within human-populated environments while adhering to social norms, is a fundamental yet challenging task in robotics \cite{coreChallengesSocialRobotNavigation, francis2023principlesguidelinesevaluatingsocial}. As shown in Figure \ref{fig:scenario_example}, navigating through social navigation scenarios requires robots to interpret human intentions, adhere to social norms, and reason about spatial and temporal interactions to respond to dynamic environments. While promising, learning-based methods that are trained on small datasets and conventional methods are often validated in controlled scenarios with a small number of people, thus falling short in handling the complexity and nuance in dynamic real-world social navigation scenarios~\cite{coreChallengesSocialRobotNavigation,rethinking_social_navigation}.

Recently, the research community has begun to explore whether advances in large Vision-Language Models (VLMs) can be leveraged as part of a solution to social robot navigation, as they have demonstrated strong capabilities in contextual understanding, commonsense reasoning, and chain-of-thought reasoning \cite{llava, gpt4o, gemini}. Trained in diverse large-scale multimodal datasets that span various real-world scenarios, large VLMs often learn underlying patterns of human behavior that may implicitly encode an understanding of social norms \cite{hu2024vivabenchmarkvisiongroundeddecisionmaking}. 
However, in social navigation, the scene understanding capabilities of VLMs remains underexplored:
Recent works like VLM-Social-Nav~\cite{vlmSocialNav} have shown that using large VLMs for social robot navigation is promising, but their evaluations are limited to a small number of controlled scenarios and offer only preliminary insights. Moreover, studies such as SPACE~\cite{spatialCognitionFrontierModels} indicate that state-of-the-art large VLMs still lack robust spatial reasoning, raising questions about whether VLMs can understand scenes of complex, realistic social navigation scenarios at all or propose socially compliant actions for robots. 

In light of these limitations, it remains essential to systematically evaluate whether VLMs can robustly handle what we consider as three critical dimensions of social robot navigation: \textbf{(1)} spatial reasoning \cite{kessler2024human}, \textbf{(2)} spatiotemporal reasoning \cite{EKSTROM20231037}, and \textbf{(3)} the ability to  interpret complex human intentions \cite{moussaid2010walking, moussaid2009experimental}. 
Existing evaluations have offered only partial assessments \cite{vlmSocialNav, spatialCognitionFrontierModels}, often focusing on controlled settings or lacking temporal components, leading to an incomplete picture of how effectively large VLMs can infer human intentions and comply with social norms in realistic, dynamic scenarios. 
This gap underscores the need for a comprehensive benchmark that rigorously tests these capabilities and may guide the development of VLMs tailored to social robot navigation.

\begin{figure*}[t]
    \centering
    \includegraphics[width=0.9\textwidth]{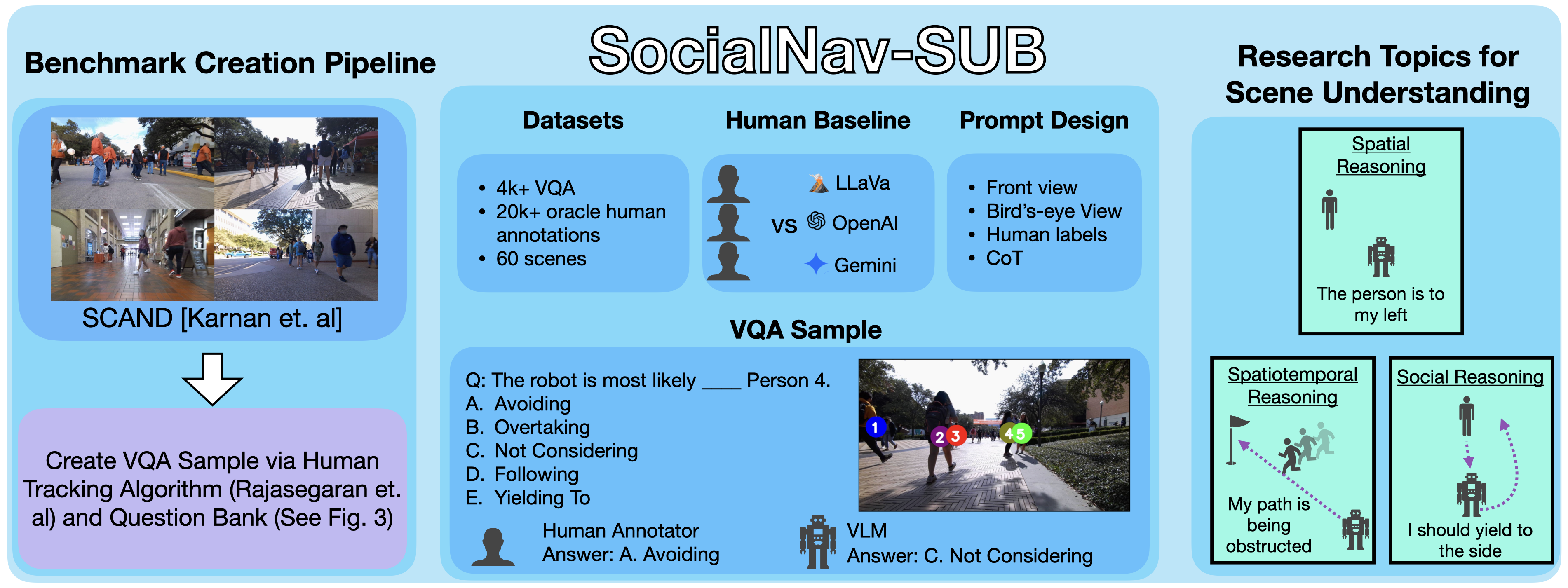}
    \caption{\textbf{An overview of {\name}}, which facilitates the systematic evaluation of VLMs in social robot navigation scenarios. Using SCAND data, human-labeled VQA datasets, and various VLMs, this framework offers the evaluation of VLMs across multiple dimensions of scene understanding for social robot navigation that can enable advancements in prompt designs, social reasoning, and social robot navigation research in general.}
    \vspace{-8pt}
    \label{fig:overview}
\end{figure*}

In this paper, we introduce the Social Navigation Scene Understanding Benchmark ({\name}), a novel Visual Question Answering (VQA) benchmark designed to evaluate VLMs on social robot navigation tasks. Shown in Figure \ref{fig:overview}, our benchmark utilizes data from a human-subject study conducted using social navigation scenarios from the SCAND dataset~\cite{scand, scanddataset}, a robot social navigation dataset of socially compliant navigation demonstrations with dense crowds and diverse social settings. 
We use our comprehensive human-labeled VQA dataset to serve as ground-truth labels to systematically evaluate the performance of VLMs on scene understanding for social robot navigation for real-world scenarios. 
We run experiments on state-of-the-art large VLMs which reveal \emph{notable performance gaps between state-of-the-art large VLMs and both human and rule-based baselines.} 

SocialNav-SUB is a first-of-its-kind benchmark that enables roboticists to systematically evaluate and refine VLMs for real-world social robot navigation scenarios. By bridging the gap between VLM capabilities and the challenges of social robot navigation, our work provides a foundation for advancing the use of VLMs for social robot navigation. Our contributions are as follows:
\begin{enumerate}
    \item \textbf{Social Navigation Scene Understanding Dataset:} We provide a human-labeled VQA dataset of 4968 unique questions and the accompanied human responses (serving as ground-truth labels) for social robot navigation tasks.
    \item \textbf{Social Navigation VQA Benchmark for VLMs:} We introduce the first VQA benchmark for assessing VLMs’ capabilities in social robot navigation scenarios using 60 unique  scenarios from SCAND that evaluates agreement with human answers.
    \item \textbf{Experiments using state-of-the-art large VLMs on our benchmark:} We evaluate several large VLMs (e.g., Gemini 2.0 and 2.5 \cite{gemini}, GPT-4o \cite{gpt4o}, OpenAI o4-mini \cite{o4mini}, LLaVa-Next-Video \cite{llava-next}) on our benchmark against human and rule-based baselines. All models perform worse than human oracle and rule-based performance.
\end{enumerate}

\vspace{-5pt}
\label{sec:introduction}

\section{Related Work}
\vspace{-5pt}
{\bf VLMs in Robotics.} In robotics, VLMs have demonstrated considerable potential for various tasks such as robotic manipulation \cite{pivot}, task planning \cite{yang2024guiding}, and human-robot interaction \cite{chang2023var,chang2023data,dong2023hubo}. The success of VLMs can be attributed to their ability to associate vision and language and generalize to unseen data in a zero-shot manner. For navigation, VLMs have been used for waypoint specification \cite{pivot, convoi}, and instruction following \cite{mobilityVLA, behav, lelan}. However, these approaches often struggle in complex real-world environments, particularly in dynamic environments, due to limitations in VLMs’ spatial reasoning capabilities \cite{spatialCognitionFrontierModels, spatialvlm, sparkleSpatialCapabilities}. This gap highlights the need for specialized evaluations and improvements of VLMs for tasks in dynamic environments, especially social navigation.

{\bf Social Robot Navigation.}  
Early social robot navigation approaches relied on model-based techniques, such as the Social Force Model (SFM) \cite{helbing1995social} and proxemics-based methods \cite{proxemics}, which used hand-engineered features to plan paths for robots. Learning-based techniques for social robot navigation, including Learning from Demonstration (LfD) \cite{hirose2023sacson, scand} and Reinforcement Learning (RL) \cite{zhu2021deep,chen2017socially,chen2019crowd,liu2023intention,liu2024height}, have shown promise in enabling robots to acquire and adapt socially compliant behaviors but are often trained on small and specialized data or simulations and struggle to generalize to complex dynamic scenarios. To address this, datasets for social robot navigation \cite{scand, nguyen2023toward} have been developed to provide more diverse and realistic social navigation scenarios, which can lead to improved generalization in social navigation models \cite{sacson}. More recently, VQA datasets for social robot navigation have been explored \cite{social-llava}, but are limited to qualitative evaluation and single images, when crucial information, such as a person's trajectory, may require a video representation. Fine-tuned VLMs have been explored for social robot navigation \cite{vlmSocialNav, social-llava}, but are often evaluated in a limited number of simple, controlled scenarios. These scattered findings suggest that while VLMs \textit{may} enhance social robot navigation, the specific capabilities that drive any observed improvements have yet to be clearly identified. Our work addresses this limitation by introducing a specialized benchmark to systematically evaluate whether VLMs can effectively perform spatial reasoning, spatiotemporal reasoning, and social reasoning for numerous social navigation scenarios.

{\bf VQA Benchmarks for VLMs.} Recent years have seen the development of various VQA benchmarks to evaluate VLMs, assessing capabilities such as spatial reasoning \cite{spatialCognitionFrontierModels}, temporal reasoning for robot navigation \cite{anwar2024remembr, zhang2023robustrobot3dperception}, scene understanding for autonomous driving~\cite{wang2025embodied, sreeram2024probing}, and physical world comprehension \cite{anonymous2024physbench}. 
While these benchmarks have advanced our understanding of VLMs’ capabilities, they often lack specific focus on social robot navigation. Our work addresses this gap by introducing a specialized VQA benchmark for social robot navigation.
\label{sec:relatedwork}

\section{SocialNav-SUB}
\vspace{-5pt}
To evaluate VLMs on scene understanding for social robot navigation, we present the \textbf{Social Navigation Scene Understanding Benchmark (\name)}, a VQA benchmark for evaluating VLMs in social navigation scenarios. Following recent works that have demonstrated the effectiveness of visual grounding and object-centric representations \cite{pivot, yang2023set, wang2025embodied}, we provide numbered labels within visual markers for objects of relevance (in our case, pedestrians) for prompting and object-centric annotations; this provides the benchmarked VLMs clear visual references and contextually rich instructions. {\name} is built on top of social navigation scenarios from SCAND that provide varying levels of crowd density and social navigation interactions and features the following: \emph{Challenging social navigation scenarios} that capture the complexities of crowded and dynamic human environments; \emph{Object-centric representations} combining both the robot’s visual perspective and a bird’s-eye view (BEV) containing pedestrian coordinate tracking for a richer object-centric representation; \emph{A diverse question set} probing spatial reasoning, temporal understanding, and social reasoning; and \emph{A robust human baseline}, where multiple annotators provide ground-truth responses for each scenario.
All above features are expanded in the following subsections below.

\subsection{Challenging Social Navigation Scenarios}
\vspace{-5pt}
\label{subsec:method_scenarios}
To effectively evaluate VLMs’ scene understanding capabilities in practical social robot navigation settings, we leverage the SCAND dataset~\cite{scand} to construct {\name}. SCAND features social robot navigation data collected by teleoperated mobile robots navigating in diverse and potentially crowded scenarios. In particular, we extract segments from SCAND that showcase moderate to high crowd density (average of 6.65 humans per scene, std. dev.: 2.80), close pedestrian proximity, and dynamically changing human motion. As illustrated in Figure~\ref{fig:scenario_example}, these densely occupied scenarios typically involve pedestrians that obstruct the robot's direct path to its goal. Hence, the teleoperated robots show complex, socially compliant interactions with the pedestrians, making these samples valuable for evaluating VLMs' scene understanding capabilities in real-world social navigation scenarios.

\begin{figure}[t]
    \centering
    \includegraphics[width=0.9\textwidth]{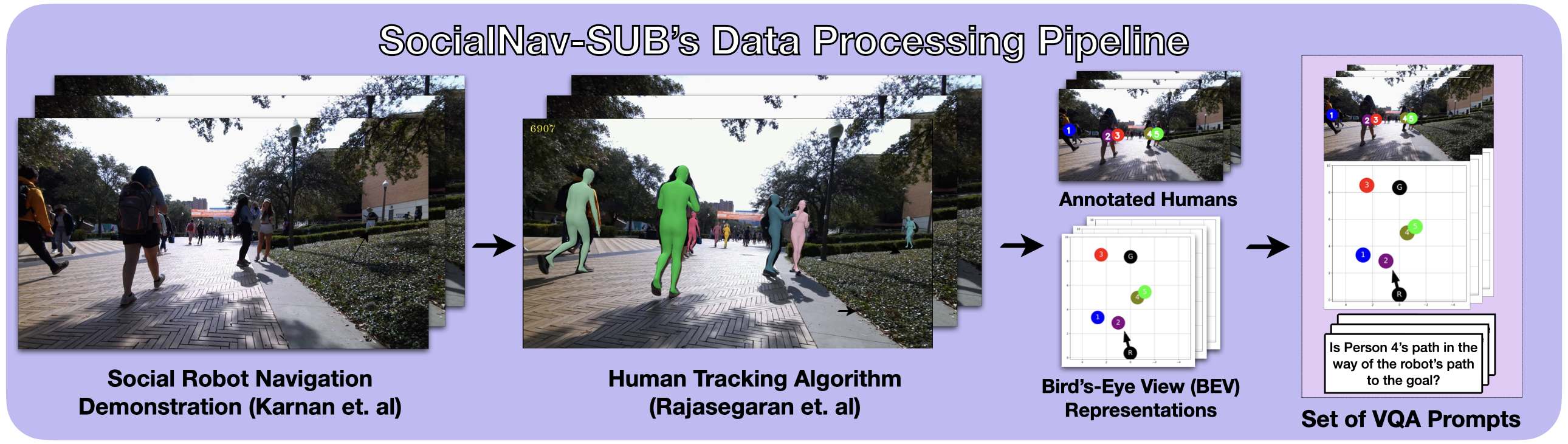}
    \caption{\textbf{The data processing pipeline for VQA prompts in {\name}.} We first mine social robot navigation scenarios from SCAND \cite{scand}, then use the PHALP algorithm \cite{phalp} to provide human tracking and estimations of 3D locations, which are used to construct BEV representations of the scene and annotated images. Along with the annotated images and BEV representations, a set of carefully designed questions (more details in Appendix \ref{appendix:questions}) that evaluate spatial reasoning, spatiotemporal reasoning, and social reasoning are used to provide VQA prompts.}
    \vspace{-8pt}
    \label{fig:data_processing_example}
\end{figure}

\subsection{Rich and Object-Centric Visual Representations}
\vspace{-5pt}
\label{subsec:method_representations}
The samples extracted from the SCAND dataset are in the form of RGB image sequences captured by the front-view camera mounted on the robot. While 2D image sequences may suffice for humans to infer the underlying spatial and social relations between the robots and pedestrians, state-of-the-art large VLMs are not necessarily good at extracting spatial or fine-grained object-level information from the same visual queries~\cite{spatialCognitionFrontierModels}. To mitigate this issue, some recent studies have shown that augmenting images with additional annotations (e.g., bounding boxes, color-coded labels) using off-the-shelf models can improve VLM performance in VQA tasks~\cite{pivot, yang2023set}. 

Building on these visual prompting insights, we augment the original data samples with additional \emph{object-centric} representations leveraging off-the-shelf vision models. As shown in Figure~\ref{fig:data_processing_example}, we begin by employing the human tracking algorithm, PHALP~\cite{phalp}, which tracks pedestrians and provides estimations of their 3D poses relative to the camera frame using monocular video input. Using the robot odometry data from SCAND, we transform the relative human poses at future timesteps into global poses relative to the robot pose in the initial frame, and apply Kalman smoothing to smooth the human poses. Afterwards, we use the camera intrinsics and extrinsics provided by SCAND to project the 3D coordinates of pedestrians into both front-view and BEV images. Finally, we annotate human positions in both views with numbered, color-coded circles. The resulting images with combined views preserve the original scene context while providing additional spatial and object-level information in a clear and structured format. In practical robotics stacks, such BEV representations can be constructed in real-time by either with learning-based methods \cite{zhang2025creste} or by utilizing tracking cameras, depth sensors, and camera matrices to estimate global positions \cite{tadic2022perspectives, aharony2024comparative}. Therefore, by querying VLMs with these enriched, object-centric visual inputs, {\name} can provide practical insights into how to best leverage and complement state-of-the-art large VLMs for practical application in social robot navigation. To ensure fair comparisons between VLMs' outputs and human responses, the same set of visual inputs are provided to human annotators.

\subsection{Diverse Scene Understanding Questions}
\vspace{-5pt}
\label{subsec:method_questions}
Following the aforementioned data processing pipeline, we construct a set of samples consisting of multi-view image sequences with object-centric annotations, each representing a $2.5$\,s segment sampled at 4\,Hz. To comprehensively evaluate VLMs’ scene understanding capabilities in social robot navigation, we design a set of multiple-choice questions (see Table \ref{appendix:questions} for more details and Appendix \ref{appendix:vqa_prompt} for an example VQA prompt) that probe across three categories: 1) \textbf{Spatial reasoning:} Questions about describing the \emph{spatial relations} in a \emph{single frame}; 2) \textbf{Spatiotemporal reasoning:} Questions about describing the \emph{motion} of the robot and pedestrians \emph{over time}; and 3) \textbf{Social reasoning:} Questions that \emph{infer whether} the robot and pedestrians are interacting and \emph{how} they interact.

These categories of questions map onto what we see as key challenges of social robot navigation: perceiving spatial relations among participants (spatial reasoning), tracking their evolution as people move (spatiotemporal reasoning), and recognizing how humans and robots interact in social navigation (social reasoning). 
By evaluating VLMs across these dimensions, we gain a fine-grained understanding of where models excel or struggle in interpreting social navigation scenes.

\subsection{Robust Human Baseline from Human-Subject Study}
\vspace{-5pt}
\label{subsec:method_humans}
We conducted human-subject studies to collect human responses as ground-truth labels for these questions under an IRB-approved protocol. Given the subjective nature of many questions, particularly those related to social reasoning, we collected responses from at least five human participants for each scenario. Participants were recruited via Prolific~\cite{prolific_website} and were asked to complete a questionnaire containing questions for multiple randomly sampled scenarios. 

By gathering this distribution of human responses, we can measure how closely each VLM output aligns with human judgments by computing the agreement between VLM answers and all human answers for a given question, which indicates the extent to which a model’s performance approaches human-level responses. We define two metrics, \textbf{Probability of Agreement (PA)} and \textbf{Consensus-Weighted PA}, to measure how closely a set of answers (from a VLM, a particular human, or a rule-based baseline) aligns with human responses overall. Let $N_Q$ be the total number of questions; $N_H$ be the number of human respondents per question; $A_q$ be the evaluated answer (from a VLM or one human) to question $q$; and $A^h_{q,i}$ be the $i$-th human's answer for question $q$, where $i \in \{1,\dots,N_H\}$.

We define \emph{Probability of Agreement (PA)} as the following:

\begin{equation}\label{eq:pa}
   \text{PA} 
   \;=\; \frac{1}{N_Q}\;\sum_{q=1}^{N_Q} 
            \Bigl(\frac{1}{N_H} \sum_{i=1}^{N_H} \mathbb{I}[\,A_q = A^h_{q,i}\,]\Bigr),
\end{equation}
where $\mathbb{I}[\cdot]$ is an indicator function which outputs 1 if $A_q$ (the evaluated answer) exactly matches the $i$-th human's response $A^h_{q,i}$, and 0 otherwise for the corresponding multiple-choice question $q$. Summing over all human responses for each question yields the fraction of total (answer, human answer) pairs that agree. PA is essentially the expected cosine similarity between the model’s predictions and the distribution of
human responses. A higher PA indicates that the evaluated answers coincide more frequently with the collected human responses. We empirically found that it is common for humans to disagree on answers, indicating there is a degree of judgement involved for particular questions. This motivates a metric that can be more forgiving for subjective questions that humans disagree on and emphasize questions that have a strong consensus, to which we establish \emph{Consensus-Weighted Probability of Agreement (CWPA)}. We start by defining
\[
   \mathrm{HA}_q 
   \;=\;
   \max_{\alpha}
   \Bigl\{
      \frac{\text{\#(humans who answered } \alpha \text{ for question }q)}{N_H}
   \Bigr\},
\]
i.e., $\mathrm{HA}_q$ is the fraction of human respondents that chose the most common answer $\alpha$ for question $q$. We then define:
\begin{equation}\label{eq:norm-pa}
   \text{CWPA}
   \;=\;\frac{1}{N_Q} 
         \sum_{q=1}^{N_Q}
         \Bigl( 
            \frac{1}{N_H \,\mathrm{HA}_q} 
            \sum_{i=1}^{N_H} 
               \mathbb{I}[\,A_q = A^h_{q,i}\,]
         \Bigr).
\end{equation}

In this formulation, each agreement with a human response for question $q$ is scaled by $1/\mathrm{HA}_q$. Consequently, questions on which humans mostly concur (i.e., high $\mathrm{HA}_q$) impose a greater penalty for incorrect answers, while questions where humans are more divided have a lower penalty. This weighting ensures that VLM (or human) answers are held to a higher standard on ``easier" questions where strong human agreement exists. Similar agreement-based metrics have been adopted to account for variability among human annotators when constructing VQA benchmarks~\cite{antol2015vqa}. Unlike their metrics, PA does not rely on a heuristically selected threshold to saturate the accuracy. CWPA further extends this by introducing a novel weighting scheme based on human consensus. 

In addition to evaluation metrics, we utilize the human responses to construct two human baselines: An \emph{Average Human Baseline}, which measures on average how often one human’s response agrees with all other human responses and serves as an indicator of average human performance but may be susceptible to noise in responses from online human participants; and A \emph{Human Oracle Baseline}, which selects the most common answer for each question from the human distribution and serves as a more robust estimate of expert-level human performance. 
\label{sec:method}

\section{Empirical Results}
\vspace{-5pt}
Our central research question examines \textit{how well state-of-the-art large VLMs that support image sequences capture spatial reasoning, scene understanding, and social reasoning in social robot navigation scenarios}. Focusing on this question, we aim to rigorously assess the capabilities and limitations of VLMs for understanding complex social robot navigation environments. We establish the benchmark with several representative models supporting video inputs across three categories: 1) {\bf closed-source, general-purpose VLMs}, including GPT-4o~\cite{gpt4o} and Gemini 2.0~\cite{gemini}, which demonstrate strong overall performance in VQA tasks; 2) {\bf reasoning VLMs}, including OpenAI o4-mini~\cite{o4mini} and Gemini 2.5~\cite{gemini}, which are  fine-tuned to enhance vision-language reasoning capabilities. While too computationally intensive for real-time deployment, they may be distilled into faster models~\cite{guo2025deepseek} suitable for robotics applications; and 3) {\bf open-source, deployable} VLMs, including LLaVa-Next-Video~\cite{llava-next}, which can run locally and are thus well-suited for robotics applications.

\subsection{Experiment Process}
\vspace{-5pt}
\label{subsec:experiments_process}
Our experiment process begins by presenting survey prompts alongside their visual and BEV representations to the VLM, using the data processing pipeline previously shown in Figure \ref{fig:data_processing_example}. The format given to the VLMs closely resembles the same visual and text format that was received by human participants, ensuring fair comparison. Furthermore, we use chain-of-thought (CoT) reasoning as a prompting technique to carry out our experiments, since this is highly similar to the sequential manner in which humans provided answer labels, allowing for fair comparison. Specifically, our usage of CoT provides the previous answers of the VLM for future questions which may help it deduce the answer to question; for example, the pedestrian is at the left in the beginning and the end and the goal is on the right, so the pedestrian is likely not obstructing the path to the goal. The responses generated by the VLM are then compared against human responses from the human dataset using the PA and CWPA metrics, previously defined in Equations \ref{eq:pa} and \ref{eq:norm-pa}

Humans can naturally infer the underlying spatial and social relations between the robots and pedestrians, making them excellent reference points of performance. On the other hand, are large VLMs truly necessary for analyzing these social robot navigation scenarios, or can a simpler, rule-based system suffice? To address both of these, we utilize the two human baselines previously defined in Section \ref{subsec:method_humans}, the \emph{Human Oracle Baseline} and the \emph{Average Human Baseline}, as well as a \emph{Rule-Based Baseline}, which uses the position data of pedestrians in the scene and uses a set of hand-crafted rules to generate answers to VQA prompts (for more details, see Appendix \ref{appendix:rule_baseline}).

\begin{table}[t]
\centering
\caption{\textbf{Average Performance Across Question Categories.} The metrics used are PA and CWPA for all questions and for each question category, along with standard error across the questions. We highlight in bold the strongest VLM PA results, which may be statistically tied. 
}
\resizebox{\textwidth}{!}{%
\begin{tabular}{c c| cc | cc | cc | cc}
\toprule
\bfseries Category & \bfseries Model 
& \multicolumn{2}{c|}{\textbf{All}}
& \multicolumn{2}{c|}{\textbf{Spatial Reasoning}}
& \multicolumn{2}{c|}{\textbf{Spatiotemporal Reasoning}}
& \multicolumn{2}{c}{\textbf{Social Reasoning}} \\
\cmidrule(lr){3-4} \cmidrule(lr){5-6} \cmidrule(lr){7-8} \cmidrule(lr){9-10}
& & PA & CWPA 
& PA & CWPA 
& PA & CWPA 
& PA & CWPA \\
\midrule

\multirow{3}{*}{\textbf{Baseline}} 
&
Human Oracle
& 0.74 ± 0.00 & 1.0 ± 0.00
& 0.71 ± 0.01 & 1.0 ± 0.00
& 0.73 ± 0.01 & 1.0 ± 0.00
& 0.76 ± 0.01 & 1.0 ± 0.00 \\
&
Average Human
& 0.60 ± 0.00 & 0.80 ± 0.00
& 0.56 ± 0.01 & 0.79 ± 0.00
& 0.59 ± 0.01 & 0.80 ± 0.00
& 0.62 ± 0.00 & 0.81 ± 0.00 \\

& Rule-Based 
& 0.64 ± 0.00 & 0.84 ± 0.00
& 0.57 ± 0.01 & 0.79 ± 0.01
& 0.62 ± 0.01 & 0.84 ± 0.01
& 0.71 ± 0.00 & 0.92 ± 0.00 \\

\midrule

\multirow{3}{*}{\textbf{VLM}} 

& Gemini 2.0 
& 0.58 ± 0.00 & 0.79 ± 0.00
& \textbf{0.55 ± 0.01} & 0.77 ± 0.01
& 0.46 ± 0.01 & 0.64 ± 0.01
& 0.63 ± 0.01 & 0.84 ± 0.01 \\

& Gemini 2.5 
& 0.54 ± 0.00 & 0.73 ± 0.01
& 0.51 ± 0.01 & 0.72 ± 0.01
& 0.52 ± 0.01 & 0.73 ± 0.01
& 0.55 ± 0.01 & 0.73 ± 0.01 \\

& GPT-4o 
& 0.50 ± 0.00 & 0.69 ± 0.01
& \textbf{0.56 ± 0.01} & 0.79 ± 0.01
& 0.51 ± 0.01 & 0.71 ± 0.01
& 0.47 ± 0.01 & 0.63 ± 0.01 \\

& o4-mini 
& \textbf{0.62 ± 0.01} & 0.82 ± 0.01
& 0.54 ± 0.01 & 0.74 ± 0.01
& \textbf{0.59 ± 0.01} & 0.79 ± 0.01
& \textbf{0.66 ± 0.01} & 0.87 ± 0.01 \\

& LLaVa-Next-Video 
& 0.46 ± 0.01 & 0.61 ± 0.01
& 0.35 ± 0.01 & 0.46 ± 0.01
& \textbf{0.58 ± 0.01} & 0.79 ± 0.01
& 0.48 ± 0.01 & 0.62 ± 0.01 \\

\bottomrule
\end{tabular}}

\vspace{-5pt}
\label{tab:main_performance}
\end{table}

\subsection{Benchmarking Results}
\vspace{-5pt}
\label{subsec:experiments_results}
We run our experiments by querying each VLM model once per unique question using default hyperparameters for each VLM. The average results over all questions and question categories is shown in Table \ref{tab:main_performance}. Among the models evaluated, OpenAI o4-mini achieves the highest overall performance, but still has a considerable gap compared to the human oracle and rule-based baselines. This performance gap suggests that state-of-the-art large VLMs are not yet fully ready for the challenges of scene understanding for social robot navigation.

When examining performance across the three question categories, models consistently lag behind the human oracle and the rule-based baseline, though the extent of the gap varies by category and perform up to par with the average human baseline. In spatial reasoning, the consensus among humans (human oracle) far exceeds that of the best models, indicating that current large VLMs struggle to accurately interpret spatial relationships compared to human observers. A similar finding is observed in spatiotemporal reasoning, where models show greater difficulty at capturing dynamic changes over time. In contrast, in social reasoning tasks, models perform relatively closer to human oracle levels and can even slightly outperform the average human baseline, suggesting that large VLMs are somewhat more adept at interpreting social cues and interactions than they are at understanding spatial relationships, although there remains a noticeable gap. Empirically, we found many cases of VLMs failing on questions with high human consensus in all three reasoning categories, especially in cases of high crowd densities, we provide qualitative examples within Appendix \ref{appendix:experiment_qual}.

\subsection{Discussion}
\vspace{-5pt}
Overall, our evaluation reveals that while state-of-the-art large VLMs like OpenAI o4-mini and Gemini 2.0 show promising advances, they still fall short of human oracle and rule-based performance across key reasoning tasks. Although models come closer to human oracle performance in social reasoning tasks, the results suggest that significant improvements are needed before these large VLMs can reliably support complex, real-world social robot navigation.

We also performed a series of ablation experiments to study the impact of querying strategies to the model performance. The results are summarized in Table~\ref{tab:ablation_1} (more details in Appendix \ref{appendix:ablations}). Our first ablation experiment analyzed the impact of CoT reasoning and found that it significantly enhances social reasoning performance for all models, likely due to the structured inference it provides for complex tasks. We also performed another ablation experiment investigating the impact of BEV scene representations and found that some models may benefit significantly, while other models show minimal changes. This suggests that BEV effectiveness depends on the VLM, but can be validated through {\name}. A further ablation experiment looked at the effectiveness of better spatial and spatiotemporal reasoning capabilities and found stronger performance on social reasoning questions, suggesting that current VLMs are limited by spatial reasoning capabilities but may be improved with fine-tuning on spatial reasoning data \cite{spatialrgpt, spatialvlm} while maintaining performance on higher-level scene understanding. These experiments highlight the usefulness of {\name} in informing how VLMs can be best utilized and further improved for social robot navigation.

\begin{table}[t]
\centering
\caption{\textbf{Ablation experiment of querying strategies}. The metric used is Probability of Agreement (PA). The baseline row BEV+CoT represents the performance with both CoT and BEV prompts enabled. The subsequent rows show the effects of removing either CoT or BEV components.} 
\begin{tabular}{l l c c c}
\toprule

\multirow{2}{*}{\textbf{Model}} & \multirow{2}{*}{\textbf{Ablation}} & \textbf{Spatial} & \textbf{Spatiotemporal} & \textbf{Social} \\ 
 & &\textbf{Reasoning} &  \textbf{Reasoning}&\textbf{Reasoning} \\ [0.35ex]
\hline
\noalign{\vskip 4pt}
\multirow{3}{*}{GPT-4o} & CoT+BEV & 0.56 ± 0.01 & 0.51 ± 0.01 & 0.47 ± 0.01 \\ [0.35ex]
 & No CoT  & 0.58 ± 0.01 & 0.53 ± 0.01 & 0.35 ± 0.01 \\ [0.35ex]
 & No BEV & 0.51 ± 0.01 & 0.44 ± 0.01 & 0.42 ± 0.01 \\ [0.35ex]
 \hline
 \noalign{\vskip 4pt}
 \multirow{3}{*}{LLaVa-Next-Video} & CoT+BEV & 0.35 ± 0.01 & 0.58 ± 0.01 & 0.48 ± 0.01 \\ [0.35ex]
 & No CoT  & 0.35 ± 0.01 & 0.58 ± 0.01 & 0.38 ± 0.01 \\ [0.35ex]
 & No BEV & 0.35 ± 0.01 & 0.61 ± 0.01 & 0.46 ± 0.01 \\ [0.35ex]
  \hline
  \noalign{\vskip 4pt}
 \multirow{3}{*}{Gemini 2.0} & CoT+BEV & 0.55 ± 0.01 & 0.46 ± 0.01 & 0.63 ± 0.01 \\ [0.35ex]
 & No CoT  & 0.56 ± 0.01 & 0.48 ± 0.01 & 0.58 ± 0.01 \\ [0.35ex]
 & No BEV & 0.56 ± 0.01 & 0.46 ± 0.01 & 0.64 ± 0.01 \\ [0.35ex]
\bottomrule
\end{tabular}

\label{tab:ablation_1}
\end{table}

Finally, we revisited our original assumption described in Section \ref{sec:introduction} that accurate scene understanding is a prerequisite for the practical usage of VLMs in real-world navigation tasks. To validate this claim, we carried out an experiment to examine the impact of scene understanding on the task of waypoint selection (see Appendix \ref{appendix:waypoint_selection}). Results indicated that providing additional scene context improved the alignment of answers with those chosen by human operators across all models, especially for reasoning models. These findings reinforce the value of our {\name} benchmark in advancing VLMs for real-world social robot navigation tasks. 
\label{sec:experiments}

\vspace{-5pt}
\section{Conclusion} 
\vspace{-5pt}
This paper introduced the Social Navigation Scene Understanding Benchmark (\name), a novel VQA benchmark designed to evaluate VLMs within complex social robot navigation scenarios. Drawing on crowded and dynamic environments from the SCAND dataset, {\name} provides object-centric visual representations, including augmented front-view images and BEV prompts, paired with a diverse set of questions targeting spatial, spatiotemporal, and social reasoning. By grounding these evaluations with a human-subject study, the benchmark offers clear, quantifiable metrics that reflect human-like understanding and decision-making in social navigation. {\name} advances the state of the art by highlighting specific strengths and weaknesses of current VLMs in handling intricate social scenes, thereby setting a clear agenda for future research. It enables researchers to systematically compare models, refine prompting strategies, and develop new methods to bridge the gap between machine and human understanding of social navigation scenes and allows for the iterative improvement of VLMs in real-world applications, ultimately guiding the development of more socially aware and reliable robotic systems.
\label{sec:conclusion}

\section{Limitations and Future Work} 
\vspace{-5pt}
While {\name} advances the evaluation of VLMs for social robot navigation, it has two limitations. First, the benchmark currently relies on scenarios from the SCAND dataset, which, despite the diverse scenarios and dense crowds (examples can be seen in the Appendix), is limited to social navigation in a university campus setting. Second, while initial experiments provide valuable insights, they are based on a limited set of models and scenarios; further exploration with a broader range of large VLMs, datasets, and refined methodologies is necessary to overcome these challenges and enhance the benchmark’s applicability.
\label{sec:limitations}

Looking ahead, several promising avenues can further enhance and leverage the capabilities of {\name}. First, expanding the dataset to include additional social robot navigation datasets could expand its diversity and robustness, offering a more comprehensive evaluation of model capabilities. Additionally, fine-tuning VLMs on the human dataset provided in {\name} may lead to VLMs that are more capable of social robot navigation. Another promising avenue is expanding upon the VLM models evaluated; some VLMs of interest include VLMs fine-tuned for spatial reasoning and VLMs fine-tuned for social robot navigation. Lastly, an interesting future direction is evaluating hybrid approaches that utilize VLMs in specific ways (such as social reasoning) while having dedicated modules to cover their weaknesses. By offering a targeted evaluation framework across multiple reasoning categories, {\name} can not only systematically evaluate VLM performance and highlight weaknesses but also guide future improvements in VLMs for both scene understanding and socially compliant navigation, enabling the development of more reliable real-world robotics systems. Since we will open-source {\name} and plan to reliably maintain it, much of the infrastructure and support to pursue these future endeavors will be readily available.
\label{sec:futureWork}

\vspace{-8pt}
\acknowledgments{
This work has taken place in the Learning Agents Research
Group (LARG) at UT Austin. LARG research is supported in part by NSF
(FAIN-2019844, NRT-2125858), ONR (N00014-24-1-2550), ARO (W911NF-17-2-0181, W911NF-23-2-0004, W911NF-25-1-0065), DARPA
(Cooperative Agreement HR00112520004 on Ad Hoc Teamwork) Lockheed
Martin, and UT Austin's Good Systems grand challenge.  Peter Stone
serves as the Chief Scientist of Sony AI and receives financial
compensation for that role.  The terms of this arrangement have been
reviewed and approved by the University of Texas at Austin in
accordance with its policy on objectivity in research.
We thank Georgios Pavlakos for helpful discussions on PHALP-based 3D position estimation and Haresh Karnan for clarifications on the SCAND dataset.
}

\bibliography{citations}

\newpage
\section{Appendix} 
\subsection{Waypoint Selection Experiments}
\label{appendix:waypoint_selection}
\begin{figure}[!htbp]
    \includegraphics[width=0.99\linewidth]{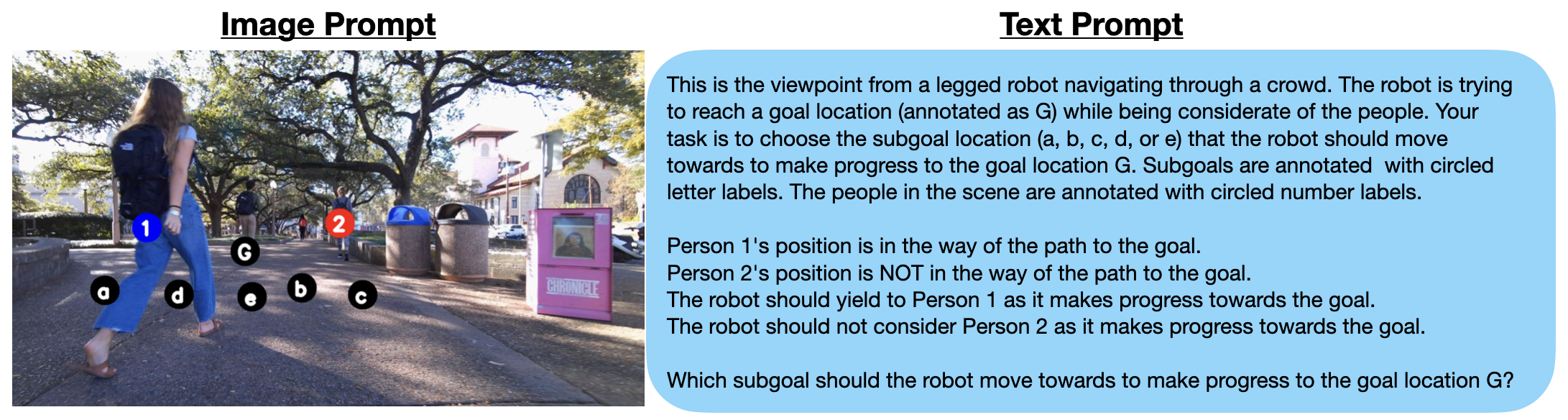}
    \caption{\textbf{An example of the waypoint selection VQA task.} This particular example highlights using scene context from the human oracle. Having no context removes the middle portion of the text prompt that includes the context, and having random context randomizes each relational action for the context (such as ``avoiding").}
    \label{fig:vqa_example_wp}
\end{figure}

To further demonstrate the practical value of {\name} in real-world social robot navigation, we conduct preliminary experiments examining how scene understanding influences VLMs' performance in waypoint selection~\cite{pivot, yang2023set}. Specifically, given visual observations and a set of candidate future waypoints annotated on the images, we prompt VLMs to select the waypoint that makes progress towards the goal while being considerate of the humans in the scene (see Figure~\ref{fig:vqa_example_wp}). One of the candidate waypoints corresponds to the ground-truth future  position of the robot as determined by the human operator. We evaluate the VLMs by comparing their selections to those made by the human operator. In addition to visual input, we incorporate scene context from various sources into the text prompts to assess their impact on waypoint selection. In this preliminary study, we condition the prompts on spatial reasoning and social reasoning context derived from predicted interactions among agents in the scene. These are provided in the form of answers to the {\bf Person End Goal Obstruction} and {\bf Robot Action to Person at End}.

The experimental results are presented in Table~\ref{tab:scene_context_comparison}. Overall, when scene context is extracted from the human oracle’s responses, VLM performance significantly improves compared to using no context or randomly generated context, and also shows slight improvement over using scene context predicted by the model itself. While preliminary, these findings suggest that accurate social scene context helps VLMs infer the ground-truth waypoints more effectively. This implies that enhancing a VLM’s scene understanding capabilities can enable it to more accurately interpret social context and subsequently select appropriate navigation actions, thereby improving overall navigation performance. Our {\name} benchmark provides the community with a valuable dataset and evaluation toolkit to support exploration along this direction. 

\begin{table}[htbp]
\centering
\caption{Accuracy of various VLMs in selecting the same waypoint as the human operator under social scene contexts from different sources: a random generator, the model itself, or the consensus from human participants (i.e., human oracle). The evaluation results are averaged over 5 runs, and we report mean accuracy ${\pm}$ standard error.}
\label{tab:scene_context_comparison}
\begin{tabular}{lcccc}
\toprule
\textbf{Model} & \textbf{No Context} & \textbf{Random} & \textbf{Same-Model} & \textbf{Human Oracle} \\
\midrule
o4-mini    & 36.14\% $\pm$ 1.31\% & 30.88\% $\pm$ 2.12\% & 38.95\% $\pm$ 1.51\% & 46.32\% $\pm$ 1.19\% \\
Gemini 2.0 & 37.19\% $\pm$ 4.75\% & 34.74\% $\pm$ 4.09\% & 41.05\% $\pm$ 2.58\% & 46.67\% $\pm$ 3.62\% \\
Gemini 2.5 & 34.39\% $\pm$ 1.97\% & 32.28\% $\pm$ 1.72\% & 37.19\% $\pm$ 1.70\% & 42.11\% $\pm$ 2.88\% \\
\bottomrule
\end{tabular}
\end{table}

\subsection{Selecting Challenging Scenes for \name}
\begin{figure}[!htbp]
  \centering
  \includegraphics[width=\linewidth]{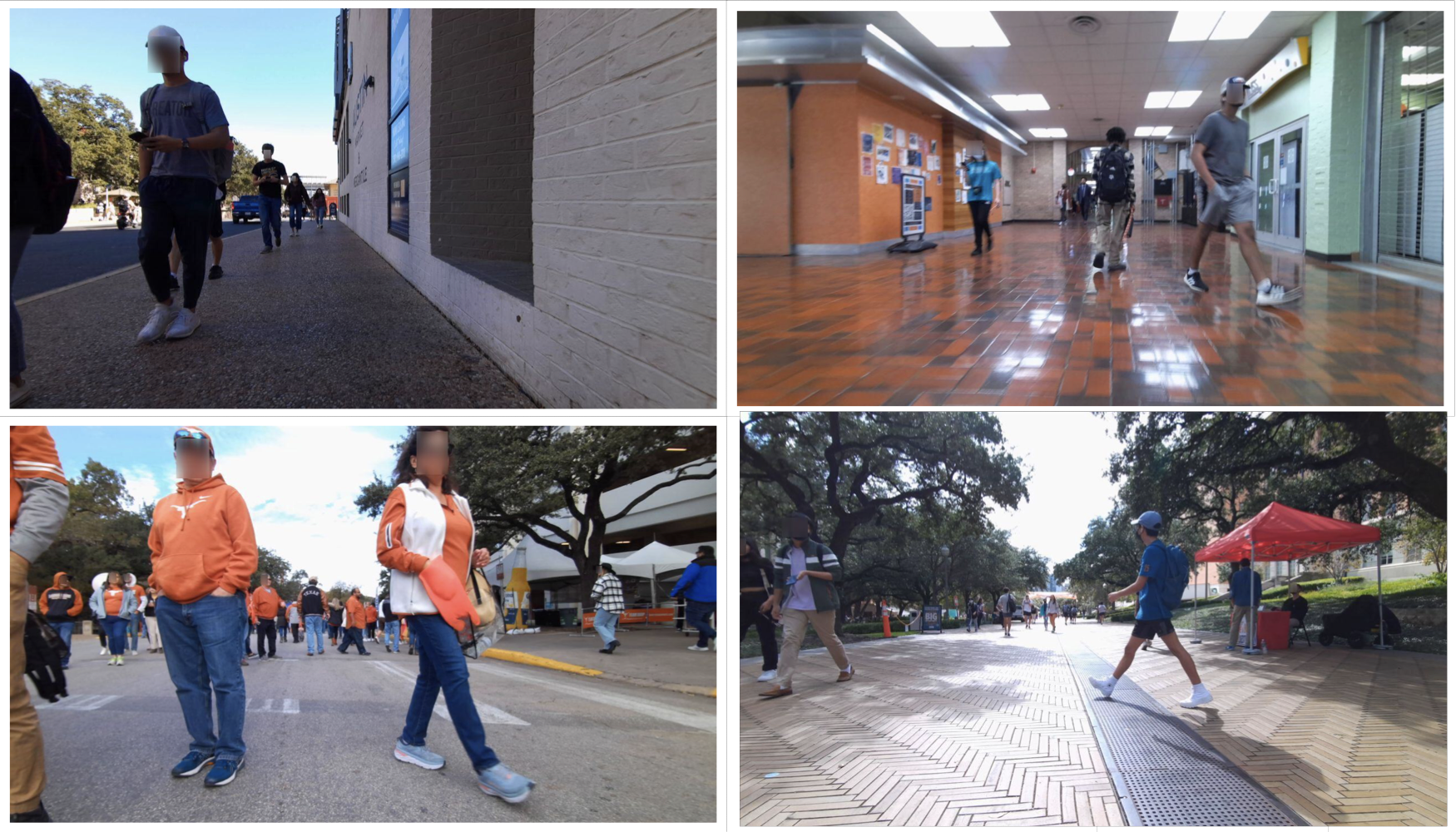}
  \caption{\textbf{Examples of scenes from \name.} These illustrate variation in environment type, crowd density, and human–robot proximity. {\name} comprises 60 social robot navigation scenarios in total.}
  \label{fig:scenes-examples}
\end{figure}

We curated 60 challenging scenes from SCAND to construct \name. Candidate scenarios were ranked using a weighted linear score over features we hypothesized to correlate with difficulty for social robot navigation: (i) crowd size, (ii) the number of people within close proximity to the robot, and (iii) the robot’s lateral (left–right) movement. We computed a weighted sum of these features and selected top-scoring scenes. The resulting set of scenarios spans across various environment types (e.g., outdoor walkways, narrow doorways/corridors, sidewalks, and street crossings) and a wide range of crowd densities (i.e., 1–13 pedestrians with mean = 6.65, SD = 2.80). Figure~\ref{fig:scenes-examples} shows four representative examples.

\subsection{Validation of 3D Pose Estimation Pipeline}
\begin{figure}[!htbp]
  \centering
  \includegraphics[width=\linewidth]{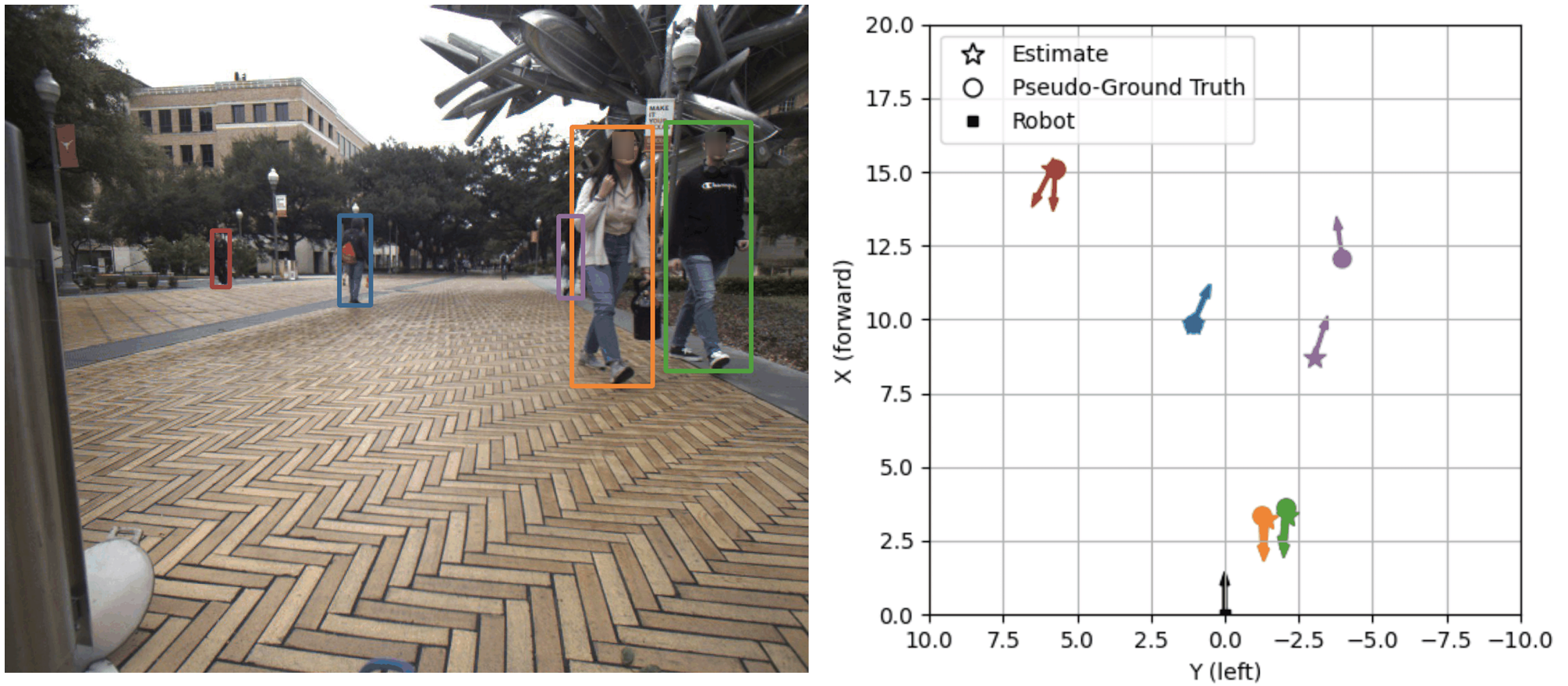}
  \caption{\textbf{CODa example for 3D pose pipeline validation.} \textbf{Left:} input image with PHALP bounding box detections. \textbf{Right:} BEV positions and headings after Kalman smoothing. Estimates are generally close (\,$<\!1$\,m displacement error) to pseudo–ground truth for well-observed pedestrians; errors increase for heavily occluded subjects.}
  \label{fig:coda_example}
\end{figure}

As described in Section~\ref{subsec:method_representations}, we estimate 3D human pose trajectories from videos using PHALP and apply Kalman smoothing to filter the estimated trajectories. Since SCAND does not provide 3D human pose labels, we validated this pipeline and tuned the hyperparameters on the CODa dataset~\cite{zhang2023robustrobot3dperception}, which provides high-quality labels 3D human pose annotations derived from human-annotated 3D bounding boxes and human-in-the-loop SLAM-based localization. We tuned the Kalman smoothing hyperparameters on CODa by minimizing a weighted sum of average displacement error and angular displacement error over trajectories across multiple scenarios. The resulting hyperparameters are then used in the \name\ pipeline. Figure~\ref{fig:coda_example} shows a CODa scenario with our estimates and the labels provided by CODa. Estimates achieve an average displacement error of $0.67 \pm 0.14\,\mathrm{m}$ across all samples. Empirically we have observed errors are lower for well-observed pedestrians and larger under heavy occlusion.

\subsection{Human-Subject Study Details}
\label{appendix:human_study}
\begin{figure}[!htbp]
    \includegraphics[width=1.0\linewidth]{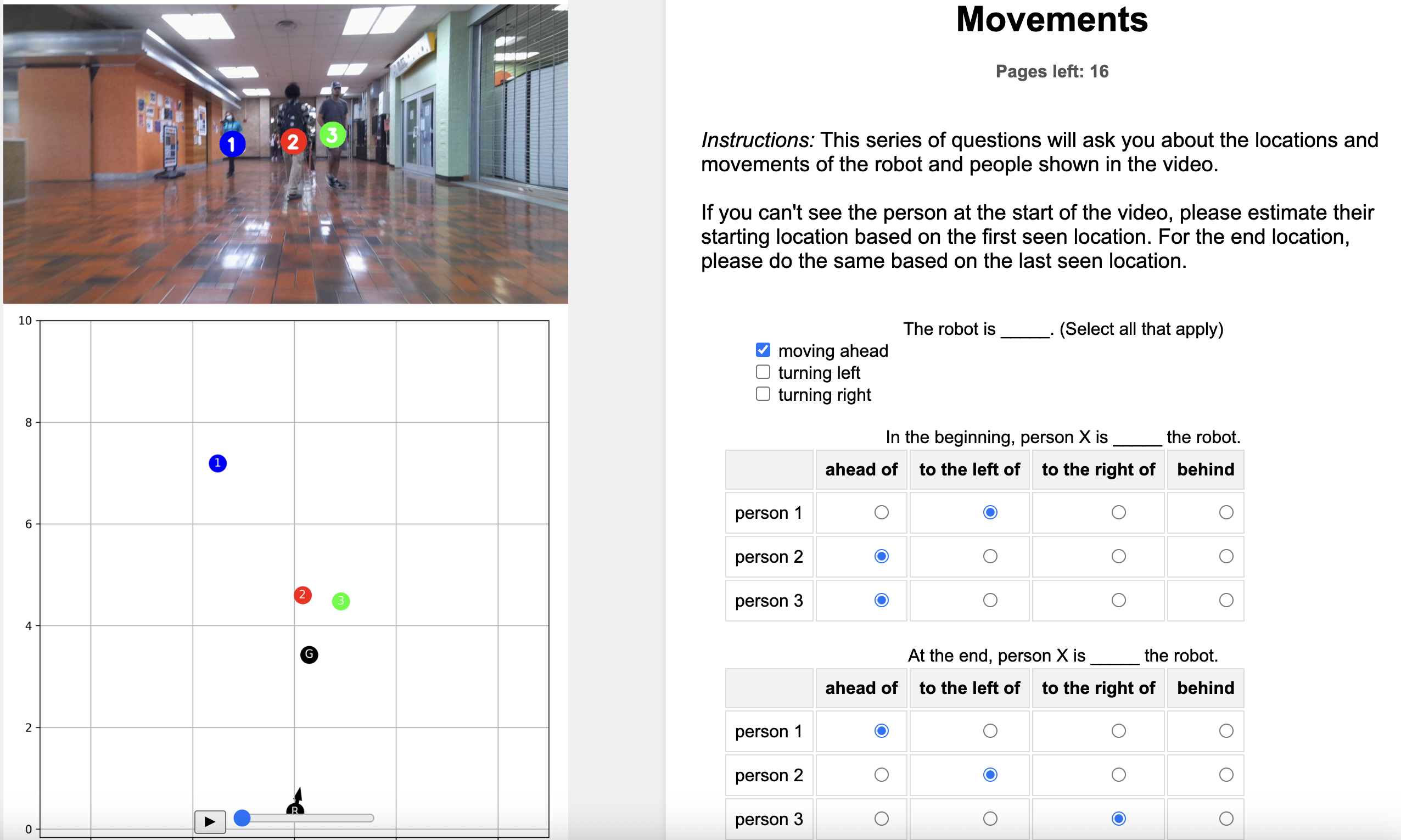}
    \caption{\textbf{An example of a survey page shown to human participants.} Prior to answering the survey questions, human subjects were given human-subject study participation instructions, requirements, and instructions about the survey content.}
    \label{fig:site_example}
\end{figure}

As mentioned in Section \ref{subsec:method_humans}, we conducted a human-subject study under an IRB-approved protocol to collect human data to establish an evaluation method for \name. We conducted our human-subject study using Prolific \cite{prolific_website} with 153 participants that were randomly selected across the U.S whose age's range from 18 to 80 (avg. 37.70, std. dev 13.40) with gender ratios of 44\% male, 54\% female, and 2\% other. Figure \ref{fig:site_example} shows an example of the interface the humans were provided for the human-subject study. Humans sequentially answered questions for each scenario in the following order: spatial reasoning questions, spatiotemporal reasoning questions, and social reasoning questions.

\subsection{VQA Prompt Details}
\label{appendix:vqa_prompt}
\begin{figure}[!htbp]
    \includegraphics[width=1.0\linewidth]{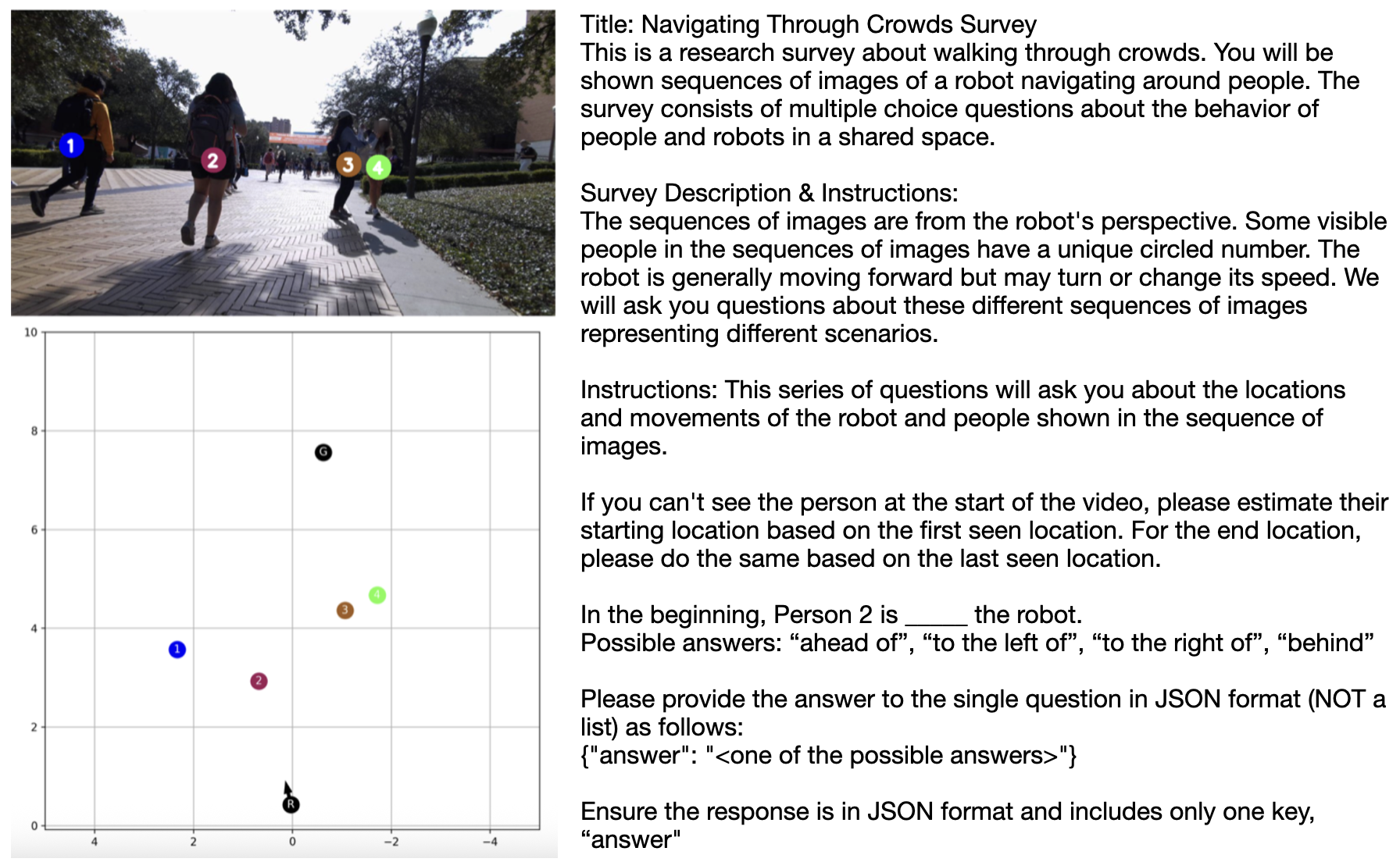}
    \caption{\textbf{An example of a full VQA prompt shown to VLMs.} This context closely resembles the instructions that were provided to human participants for the human-subject study. In addition to the image shown on the left, the VLM also receives the next 9 images in the sequence.}
    \label{fig:vqa_example_1}
\end{figure}

To provide fair comparison between humans and VLMs, we provided VLMs with highly similar input. In Figure \ref{fig:vqa_example_1}, we provide a full VQA example of what the VLM receives as input. Chain-of-thought reasoning was used in the main experiments outlined in Section \ref{sec:experiments} and this particular usage consisted of sequentially asking the VLM questions, where later questions require higher-level reasoning, and providing the VLM its answers to the relevant questions within the prompt.

\subsection{Failure Case Analysis}
\label{appendix:experiment_qual}
\begin{figure}[!htbp]
    \includegraphics[width=1.0\linewidth]{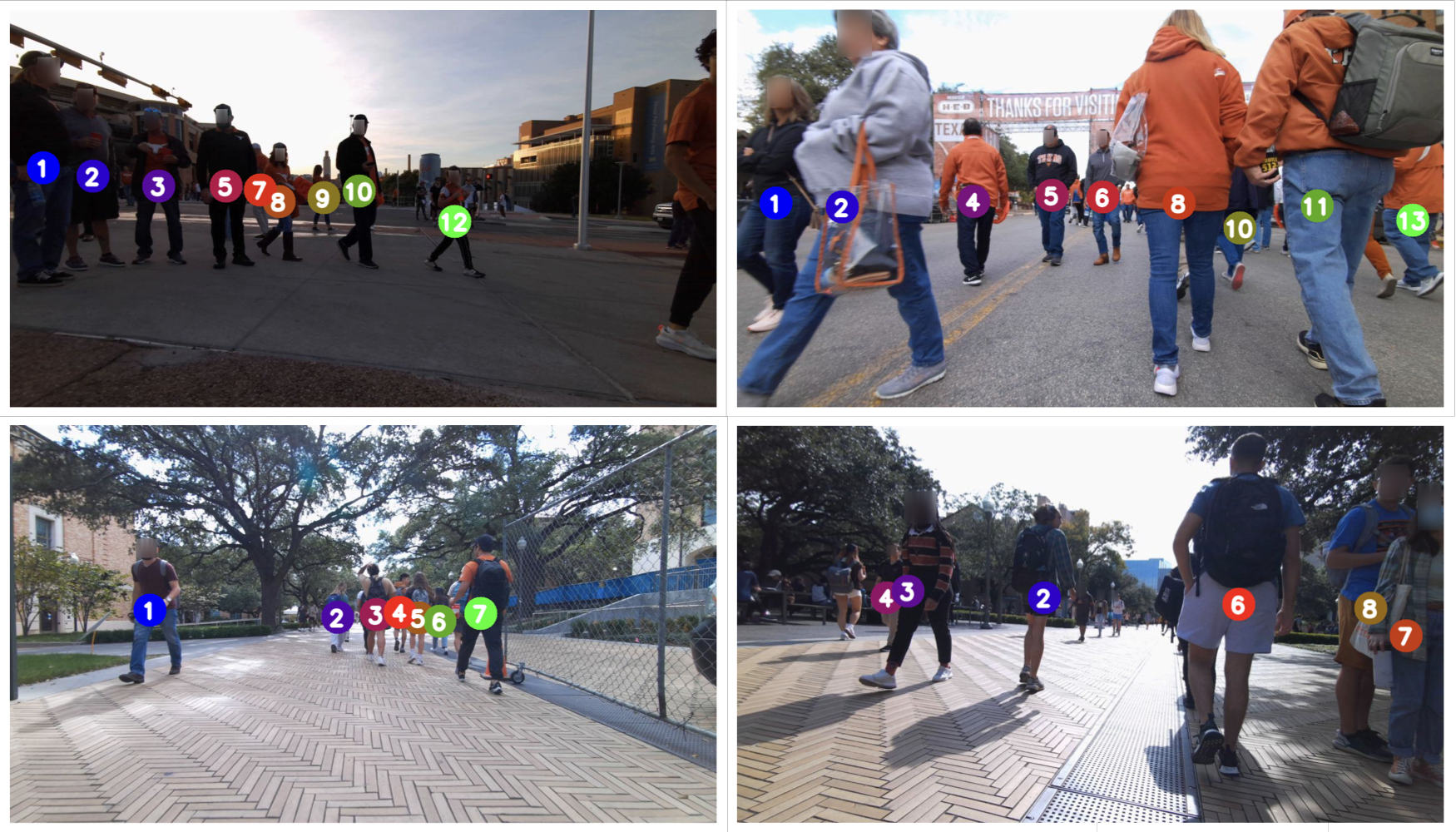}
    \caption{\textbf{Examples of failure cases for VLMs.} \emph{Top-left:} Failing to recognize that person 5 is on the left. \emph{Top-right:} Failing to recognize that person 4 ends up further away. \emph{Bottom-left:} Answering that the distant person 3 should be avoided. \emph{Bottom-right:} Incorrectly answering that an action should be taken with respect to person 7, although all humans did not think they were relevant. These examples were selected automatically based upon the entropy of the VLM answers and human answers.}
    \label{fig:fails}
\end{figure}

\begin{figure}[!htbp]
    \includegraphics[width=1.0\linewidth]{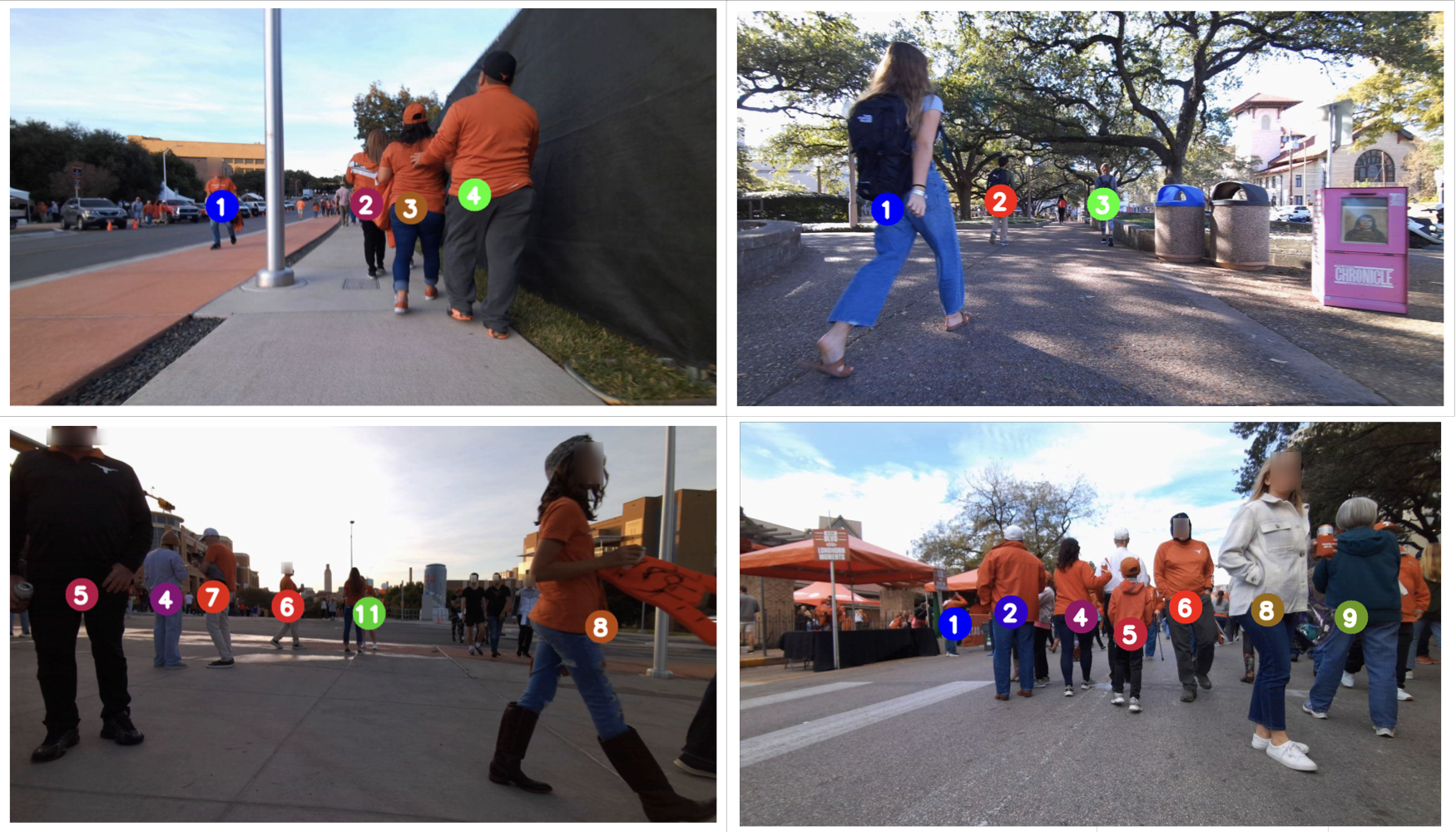}
    \caption{\textbf{Examples of success cases for VLMs.} \emph{Top-left:} All VLMs correctly infer that person 1 is not obstructing the path to the goal.  \emph{Top-right:} Gemini correctly predicts that person 1 should be avoided (the other VLMs incorrectly predict this). \emph{Bottom-left:} GPT-4o correctly answers that person 5 is on the left, whereas both Gemini and LLaVa-Next-Video answer that person 5 is behind the robot. \emph{Bottom-right:} Most VLMs (but not all) predict that person 6 is being considered as the robot is moving towards the goal, similar to the distribution among human answers. These examples were automatically selected based upon the entropy of the VLM answers and the human answers.}
    \label{fig:successes}
\end{figure}

As mentioned in Section \ref{subsec:experiments_results}, we found cases of VLMs in the experiment failing on questions with high human consensus in all reasoning categories, especially in cases of high crowd densities; we show these failure cases in Figure \ref{fig:fails}. We also highlight cases where VLMs can provide success, shown in Figure \ref{fig:successes}. These cases were automatically selected based on the entropy of the VLM answers and human answers.

\FloatBarrier

\noindent\textbf{Quantitative summary.}
We summarize quantitative findings shown in Tables \ref{tab:env_fri_summary} and \ref{tab:act_fr_summary} for VLM failure cases (defined as the model’s chosen answer received zero human probability); we report failure rates in percent and means $\pm$ standard errors. \emph{Robot Action to Person} denotes the model’s classification of the robot’s action relative to a human (e.g., \textit{yielding to}, \textit{overtaking}, \textit{avoiding}, \textit{following}, \textit{not considering}).

\begin{table}[!htbp]
\centering
\footnotesize
\setlength{\tabcolsep}{2pt}
\newcommand{\fr}[2]{#1 {\scriptsize$\pm$ #2}}
\newcommand{\fri}[3]{#1$\times$ ({\scriptsize #2 $\pm$ #3})}
\caption{\textbf{Environment summary across VLMs.} \emph{Overall FR} is the model’s failure rate with standard error in smaller type. Environment cells show \emph{failure rate increase} (FRI) relative to the model’s overall FR, with raw FR $\pm$ SE in smaller parentheses. FRI $=1$ equals overall; $>1$ is worse-than-average; $<1$ is better-than-average for that model.}
\label{tab:env_fri_summary}
\begin{tabular}{lcccc}
\toprule
Model & Overall FR & Indoors & Outdoors & Blind corner \\
\midrule
o4-mini     & \fr{6.42\%}{0.33\%}  & \fri{0.87}{5.58\%}{0.90\%}  & \fri{1.02}{6.53\%}{0.36\%}  & \fri{1.45}{9.28\%}{1.56\%} \\
GPT-4o      & \fr{23.63\%}{0.60\%} & \fri{1.39}{32.91\%}{2.15\%} & \fri{0.96}{22.65\%}{0.62\%} & \fri{1.13}{26.67\%}{2.95\%} \\
LLaVa-Next  & \fr{33.78\%}{0.67\%} & \fri{0.91}{30.82\%}{2.11\%} & \fri{1.01}{34.09\%}{0.71\%} & \fri{0.99}{33.33\%}{3.14\%} \\
Gemini 2.0  & \fr{15.00\%}{0.51\%} & \fri{1.22}{18.24\%}{1.77\%} & \fri{0.98}{14.65\%}{0.53\%} & \fri{1.16}{17.33\%}{2.52\%} \\
Gemini 2.5  & \fr{9.02\%}{0.41\%}  & \fri{0.91}{8.18\%}{1.25\%}  & \fri{1.01}{9.11\%}{0.43\%}  & \fri{1.13}{10.22\%}{2.02\%} \\
\bottomrule
\end{tabular}
\end{table}

\begin{table}[!htbp]
\centering
\footnotesize
\setlength{\tabcolsep}{3pt}
\caption{\textbf{Robot Action to Person.} Cells report failure rate (FR); parentheses show the occurrence (\%) of the VLM choosing that action. “—” denotes the action was never chosen.}
\label{tab:act_fr_summary}
\begin{tabular}{lccccc}
\toprule
Model & Yielding to & Overtaking & Following & Avoiding & Not Considering \\
\midrule
o4-mini     & 57.14\% (3.2\%)  & 50.00\% (1.8\%)  & 31.25\% (3.6\%)  & 20.59\% (7.7\%)  & 2.70\% (83.7\%) \\
GPT-4o      & 84.85\% (8.3\%)  & 80.00\% (1.3\%)  & 20.00\% (1.3\%)  & 48.84\% (53.9\%) & 1.42\% (35.3\%) \\
LLaVa-Next  & —                 & —                 & —                 & 53.13\% (88.2\%) & 6.38\% (11.8\%) \\
Gemini 2.0  & 66.67\% (0.8\%)   & 100.00\% (0.5\%)  & —                 & 46.62\% (33.3\%) & 3.83\% (65.4\%) \\
Gemini 2.5  & 62.50\% (2.0\%)   & 47.37\% (9.5\%)   & 0.00\% (2.8\%)    & 15.56\% (33.8\%) & 0.97\% (51.9\%) \\
\bottomrule
\end{tabular}
\end{table}

\paragraph{GPT-4o.}
Overall 24\% failure rate; higher indoors than outdoors (32.9\% vs 22.6\%) and at blind corners (26.7\% vs 23.5\%); for Robot Action to Person: \textit{yielding to} 85\%, \textit{overtaking} 80\%, \textit{avoiding} 48.8\%, \textit{not considering} 1.4\%.

\paragraph{o4-mini.}
Overall 6.4\% failure rate; higher at blind corners (9.3\% vs 6.2\%); Robot Action to Person: \textit{yielding to} 57\%, \textit{overtaking} 50\%, \textit{following} 31\%, \textit{avoiding} 21\%, \textit{not considering} 2.7\%.

\paragraph{LLaVa-Next.}
Overall 33.8\% failure rate; very limited Robot Action to Person diversity—only \textit{avoiding} (53.1\%) and \textit{not considering} (6.38\%).

\paragraph{Gemini 2.0.}
Overall 15.0\% failure rate; higher at blind corners (17.3\% vs 14.9\%); failure cases show more people in scene (12.26 vs 11.91); Robot Action to Person: \textit{avoiding} 46.6\%, \textit{not considering} 3.83\%, with rare but often wrong \textit{yielding to} (66.7\%) and \textit{overtaking} (100\%).

\paragraph{Gemini 2.5.}
Overall 9.0\% failure rate; Robot Action to Person shows higher action diversity but some labels remain difficult: \textit{yielding to} 62.5\%, \textit{overtaking} 47.4\%, versus \textit{avoiding} 15.6\%, \textit{not considering} 1.0\%, \textit{following} 0\%.

\FloatBarrier

\subsection{Survey Question Details}
\label{appendix:questions}

Here we show the details and qualitative descriptions of questions used throughout the benchmark by providing a question for each VQA prompt, shown in Table \ref{table:questions}. We categorize these questions according to their reasoning capability.

\begin{table}[htbp!]
\centering
\caption{\textbf{Qualitative descriptions of the text components for questions used in {\name}}, their pertaining primary reasoning capability, and the number of unique questions through {\name}. All questions are multiple-choice questions, with each VQA prompt providing the possible answers. An example of a VQA prompt can be found in Figure \ref{fig:overview} and a full example can be found in Appendix \ref{appendix:vqa_prompt}.}
\begin{tabularx}{\linewidth}{p{0.2\linewidth} X >{\raggedleft\arraybackslash}p{0.05\linewidth}}

\toprule

\textbf{VLM Reasoning Capability} & \textbf{Qualitative Description of Question} & \textbf{\# of Questions} \\

\midrule

\multirow{6}{*}{\textbf{Spatial}} 
& \textbf{Person Initial Position}: The position of the person at the beginning of the video. & 399 \\
& \textbf{Person Ending Position}: The position of the person at the end of the video. & 399 \\
& \textbf{Goal Initial Position}: The initial position of the goal with respect to the robot's view. & 60 \\
& \textbf{Goal End Position}: The end position of the goal with respect to the robot's view. & 60 \\
& \textbf{Person End Goal Obstruction}: Whether the person is obstructing the robot's path towards the goal at the end of the video. & 399 \\

\midrule

\multirow{5}{*}{\textbf{Spatiotemporal}} 
& \textbf{Robot Moving Direction}: The direction the robot is moving in the video. & 60 \\
& \textbf{Person Distance Change}: The relative distance change of the person to the robot from the beginning of the video to the end. & 399 \\
& \textbf{Person Goal Obstruction}: Whether the person is obstructing the robot's path towards the goal during the video. & 399 \\

\midrule

\multirow{12}{*}{\textbf{Social}}
& \textbf{Robot Affected by Person}: Whether the robot's (human operator's) actions are affected by the person. & 399 \\
& \textbf{Robot Action to Person}: The high-level relational action of the robot with respect to the person (e.g., the robot avoided person 2). & 399 \\
& \textbf{Person Affected by Robot}: Whether the robot's (human operator's) actions are affected by the person. & 399 \\
& \textbf{Person Action to Robot}: The high-level relational action of the person with respect to the robot (e.g., person 2 avoided the robot). & 399 \\
& \textbf{Robot Affected by Person at End}: Whether the robot's (human operator's) actions are affected by the person at the end of the video. & 399 \\
& \textbf{Robot Action to Person at End}: The high-level relational action of the robot with respect to the person at the end of the video. & 399 \\ 
& \textbf{Person Action to Robot at End}: The high-level relational action of the person with respect to the robot at the end of the video. & 399 \\

\bottomrule

\end{tabularx}
\label{table:questions}
\end{table}

\begingroup
  \renewcommand{\arraystretch}{1.2}
  
\begin{table}[htbp!]
\centering
\caption{\textbf{Full spatial reasoning questions in {\name}, with question type and options.} Here, {PERSON} can be any labeled person in the scene, e.g. Person 3.}
\begin{tabularx}{\linewidth}{p{0.5\linewidth} p{0.1\linewidth} X}
\hline
\textbf{Question} & \textbf{Type} & \textbf{Options} \\ \hline
In the beginning, \{PERSON\} is \_\_\_ the robot. 
  & Multiple Choice  
  & ahead of; to the left of; to the right of; behind \\ \hline
At the end, \{PERSON\} is \_\_\_ the robot. 
  & Multiple Choice  
  & ahead of; to the left of; to the right of; behind \\ \hline
In the beginning frame, the goal is \_\_\_ of the robot. 
  & Multiple Choice  
  & ahead; to the left; to the right \\ \hline
At the end frame, the goal is \_\_\_ of the robot. 
  & Multiple Choice  
  & ahead; to the left; to the right \\ \hline
At the end frame, is \{PERSON\}’s position in the way of the robot’s path to the goal? 
  & Multiple Choice  
  & yes; no \\ \hline
\end{tabularx}
\label{table:spatial_questions}
\end{table}

\begin{table}[htbp!]
\centering
\caption{\textbf{Full spatiotemporal reasoning questions in {\name}, with question type and options.} Here, {PERSON} can be any labeled person in the scene, e.g. Person 3. We convert the Multiple Select question into Multiple Choice by taking the power set of all options.}
\begin{tabularx}{\linewidth}{p{0.5\linewidth} p{0.1\linewidth} X}
\hline
\textbf{Question} & \textbf{Type} & \textbf{Options} \\ \hline
The robot is \_\_\_\_\_ (Select all that apply) 
  & Multiple Select  
  & moving ahead; turning left; turning right \\ \hline
At the end, \{PERSON\} ends up \_\_\_\_ the robot compared to the beginning. 
  & Multiple Choice  
  & closer to; further away from; about the same distance to \\ \hline
Is \{PERSON\}’s path in the way of the robot’s path to the goal? 
  & Multiple Choice  
  & yes; no \\ \hline
\end{tabularx}
\label{table:spatiotemporal_questions}
\end{table}

\begin{table}[htbp!]
\centering
\caption{\textbf{Full social reasoning questions in {\name}, with question type and options.} Here, {PERSON} can be any labeled person in the scene, e.g. Person 3.}
\begin{tabularx}{\linewidth}{p{0.5\linewidth} p{0.1\linewidth} X}
\hline
\textbf{Question} & \textbf{Type} & \textbf{Options} \\ \hline
Is the robot’s movement affected by \{PERSON\}? 
  & Multiple Choice  
  & yes; no \\ \hline
The robot is most likely \_\_\_ \{PERSON\}. 
  & Multiple Choice  
  & avoiding; overtaking; not considering; following; yielding to \\ \hline
Is \{PERSON\}’s movement affected by the robot? 
  & Multiple Choice  
  & yes; no \\ \hline
\{PERSON\} is most likely \_\_\_ the robot. 
  & Multiple Choice  
  & avoiding; overtaking; not considering; following; yielding to \\ \hline
In the future (after the end of the video), should the robot’s movement towards the goal be affected by \{PERSON\}? 
  & Multiple Choice  
  & yes; no \\ \hline
In the future (after the end of the video), the robot should \_\_\_ \{PERSON\} as it makes progress towards the goal. 
  & Multiple Choice  
  & avoid; overtake; not consider; follow; yield to \\ \hline
In the future (after the end of the video), \{PERSON\} will most likely \_\_\_ the robot as the robot attempts to make progress towards the goal. 
  & Multiple Choice  
  & avoid; overtake; not consider; follow; yield to \\ \hline
\end{tabularx}
\label{table:social_questions}
\end{table}

\endgroup

\FloatBarrier

\subsection{Main Experiment Question Results}
\label{appendix:additional_experiments}
We provide the question-level performance for the main experiment results from Section \ref{subsec:experiments_results} for the VLMs shown in Table \ref{tab:question_performance_vlms}, reasoning-based VLMs shown in \ref{tab:question_performance_reasoning_vlms}, and the baselines shown in Table \ref{tab:question_performance_baseline}.

\begin{table}[!htbp]
\centering
    \caption{\textbf{Performance Across Individual Questions for non-reasoning VLMs.} These results highlight the deficiencies of non-reasoning VLMs: 1) Gemini \cite{gemini} has stronger social reasoning than other non-reasoning VLMs for most questions but has worse spatial reasoning performance across most tasks compared to GPT-4o; 2) LLaVa-Next-Video \cite{llava-next} has poor spatial reasoning performance for most questions, determining the moving direction of the robot, and poor ability to infer the future action of the robot, but performs well for certain questions such as determining whether somebody is obstructing the goal and some social reasoning questions; 3) GPT-4o \cite{gpt4o} has moderate performance across tasks but lacks strong social reasoning.}
\tiny 
\begin{tabular}{llcccccc}
\toprule
\multirow{2}{*}{\textbf{Category}} 
 & \multirow{2}{*}{\textbf{Question Name}} 
 & \multicolumn{2}{c}{\textbf{Gemini 2.0}} 
 & \multicolumn{2}{c}{\textbf{GPT-4o}} 
 & \multicolumn{2}{c}{\textbf{LLaVa-Next-Video}} \\
\cmidrule(lr){3-4}
\cmidrule(lr){5-6}
\cmidrule(lr){7-8}
 & 
 & \textbf{PA} & \textbf{CW PA}
 & \textbf{PA} & \textbf{CW PA}
 & \textbf{PA} & \textbf{CW PA} \\
\midrule

\multirow{5}{*}{\textbf{Spatial}}
 & Person Initial Position      & 0.52 ± 0.01 & 0.81 ± 0.01 & 0.54 ± 0.01 & 0.84 ± 0.01 & 0.05 ± 0.00 & 0.10 ± 0.01 \\
 & Person Ending Position       & 0.38 ± 0.01 & 0.64 ± 0.02 & 0.43 ± 0.01 & 0.71 ± 0.02 & 0.24 ± 0.01 & 0.44 ± 0.02 \\
 & Goal Initial Position        & 0.69 ± 0.04 & 0.85 ± 0.04 & 0.74 ± 0.03 & 0.92 ± 0.03 & 0.14 ± 0.02 & 0.20 ± 0.04 \\
 & Goal End Position            & 0.56 ± 0.04 & 0.73 ± 0.05 & 0.65 ± 0.04 & 0.83 ± 0.04 & 0.15 ± 0.02 & 0.22 ± 0.04 \\
 & Person End Goal Obstruction  & 0.74 ± 0.01 & 0.86 ± 0.02 & 0.68 ± 0.02 & 0.78 ± 0.02 & 0.80 ± 0.01 & 0.93 ± 0.01 \\
\midrule

\multirow{3}{*}{\textbf{Spatiotemporal}}
 & Robot Moving Direction       & 0.46 ± 0.05 & 0.64 ± 0.05 & 0.57 ± 0.04 & 0.81 ± 0.04 & 0.24 ± 0.04 & 0.38 ± 0.05 \\
 & Person Distance Change       & 0.31 ± 0.01 & 0.53 ± 0.02 & 0.46 ± 0.01 & 0.74 ± 0.02 & 0.47 ± 0.01 & 0.75 ± 0.02 \\
 & Person Goal Obstruction      & 0.62 ± 0.02 & 0.76 ± 0.02 & 0.54 ± 0.02 & 0.67 ± 0.02 & 0.74 ± 0.01 & 0.89 ± 0.01 \\
\midrule

\multirow{7}{*}{\textbf{Social}}
 & Robot Affected by Person          & 0.64 ± 0.02 & 0.78 ± 0.02 & 0.50 ± 0.02 & 0.63 ± 0.02 & 0.75 ± 0.01 & 0.91 ± 0.01 \\
 & Robot Action to Person            & 0.51 ± 0.01 & 0.75 ± 0.02 & 0.37 ± 0.01 & 0.57 ± 0.02 & 0.25 ± 0.01 & 0.42 ± 0.02 \\
 & Person Affected by Robot          & 0.74 ± 0.01 & 0.88 ± 0.01 & 0.58 ± 0.02 & 0.71 ± 0.02 & 0.79 ± 0.01 & 0.94 ± 0.01 \\
 & Person Action to Robot            & 0.62 ± 0.01 & 0.86 ± 0.02 & 0.45 ± 0.02 & 0.65 ± 0.02 & 0.67 ± 0.01 & 0.92 ± 0.01 \\
 & Robot Affected by Person at end   & 0.72 ± 0.01 & 0.87 ± 0.01 & 0.55 ± 0.02 & 0.68 ± 0.02 & 0.79 ± 0.01 & 0.94 ± 0.01 \\
 & Robot Action to Person at end     & 0.60 ± 0.01 & 0.85 ± 0.02 & 0.41 ± 0.02 & 0.59 ± 0.02 & 0.08 ± 0.01 & 0.14 ± 0.01 \\
 & Person Action to Robot at end     & 0.62 ± 0.01 & 0.87 ± 0.01 & 0.40 ± 0.02 & 0.59 ± 0.02 & 0.03 ± 0.00 & 0.05 ± 0.01 \\
\bottomrule
\end{tabular}
\begin{minipage}{\textwidth}
    \centering
\label{tab:question_performance_vlms}
\end{minipage}
\end{table}

\begin{table}[!htbp]
\centering
    \caption{\textbf{Performance Across Individual Questions for Large Reasoning Models.} These results indicate that o4-mini displays worse performance across most spatial reasoning question but has strong performance on determining if a person is obstructing the path to the goal. We hypothesize, with evidence in Appendix \ref{appendix:waypoint_selection}, that better performance in these questions can result in better social reasoning performance and may be a limiting factor for o4-mini. Gemini 2.5 shows worse performance across spatiotemporal reasoning and social reasoning compared to o4-mini but comparable performance in spatial reasoning. Gemini 2.5 has a particularly difficult time in determining the moving direction of the robot compared to other models. Although we evaluated using o4-mini and Gemini 2.5 flash, we expect that these may be lower bounds on the performance for their higher-end model variations.}
\tiny 
\begin{tabular}{llcccc}
\toprule
\multirow{2}{*}{\textbf{Category}} 
 & \multirow{2}{*}{\textbf{Question Name}} 
 & \multicolumn{2}{c}{\textbf{Gemini 2.5}} 
 & \multicolumn{2}{c}{\textbf{o4-mini}}\\
\cmidrule(lr){3-4}
\cmidrule(lr){5-6}
 & 
 & \textbf{PA} & \textbf{CW PA}
 & \textbf{PA} & \textbf{CW PA} \\
\midrule

\multirow{5}{*}{\textbf{Spatial}}
 & Person Initial Position      & 0.49 ± 0.01 & 0.78 ± 0.02 & 0.49 ± 0.01 & 0.76 ± 0.02 \\
 & Person Ending Position       & 0.36 ± 0.01 & 0.59 ± 0.02 & 0.36 ± 0.01 & 0.58 ± 0.02 \\
 & Goal Initial Position        & 0.70 ± 0.04 & 0.86 ± 0.04 & 0.48 ± 0.05 & 0.58 ± 0.06 \\
 & Goal End Position            & 0.52 ± 0.04 & 0.67 ± 0.05 & 0.48 ± 0.05 & 0.60 ± 0.06 \\
 & Person End Goal Obstruction  & 0.66 ± 0.02 & 0.77 ± 0.02 & 0.81 ± 0.01 & 0.93 ± 0.01 \\
\midrule

\multirow{3}{*}{\textbf{Spatiotemporal}}
 & Robot Moving Direction       & 0.34 ± 0.04 & 0.50 ± 0.06 & 0.56 ± 0.04 & 0.80 ± 0.04 \\
 & Person Distance Change       & 0.47 ± 0.01 & 0.75 ± 0.02 & 0.45 ± 0.01 & 0.71 ± 0.02 \\
 & Person Goal Obstruction      & 0.60 ± 0.02 & 0.73 ± 0.02 & 0.73 ± 0.01 & 0.87 ± 0.01 \\
\midrule

\multirow{7}{*}{\textbf{Social}}
 & Robot Affected by Person          & 0.57 ± 0.02 & 0.71 ± 0.02 & 0.73 ± 0.01 & 0.89 ± 0.01 \\
 & Robot Action to Person            & 0.44 ± 0.02 & 0.67 ± 0.02 & 0.58 ± 0.01 & 0.84 ± 0.02 \\
 & Person Affected by Robot          & 0.70 ± 0.02 & 0.83 ± 0.02 & 0.77 ± 0.01 & 0.91 ± 0.01 \\
 & Person Action to Robot            & 0.58 ± 0.02 & 0.80 ± 0.02 & 0.60 ± 0.02 & 0.83 ± 0.02 \\
 & Robot Affected by Person at end   & 0.56 ± 0.02 & 0.68 ± 0.02 & 0.77 ± 0.01 & 0.91 ± 0.01 \\
 & Robot Action to Person at end     & 0.44 ± 0.02 & 0.62 ± 0.02 & 0.62 ± 0.01 & 0.87 ± 0.01 \\
 & Person Action to Robot at end     & 0.58 ± 0.01 & 0.81 ± 0.02 & 0.58 ± 0.02 & 0.82 ± 0.02 \\
\bottomrule
\end{tabular}
\begin{minipage}{\textwidth}
\label{tab:question_performance_reasoning_vlms}
\end{minipage}
\end{table}

\begin{table}[!htbp]
\centering
\caption{\textbf{Performance Across Individual Questions for Baselines.} For the Human Oracle and Average Human baselines, these results highlight questions that humans disagreed on more often, showing that determining spatial labels for humans was more disagreeable than social reasoning questions. The rule-based baseline performance indicates that it struggles with determining what the initial and ending position of humans are as well as determining if a person gets closer to further away, showing that it is not as trivial as determining a cutoff value for this from rules described in \ref{appendix:rule_baseline}.
}
\tiny 
\begin{tabular}{llcccccc}
\toprule
\multirow{2}{*}{\textbf{Category}} 
 & \multirow{2}{*}{\textbf{Question Name}} 
 & \multicolumn{2}{c}{\textbf{Human Oracle}} 
 & \multicolumn{2}{c}{\textbf{Average Human}} 
 & \multicolumn{2}{c}{\textbf{Rule-Based}} \\
\cmidrule(lr){3-4}
\cmidrule(lr){5-6}
\cmidrule(lr){7-8}
 & 
 & \textbf{PA} & \textbf{CW PA}
 & \textbf{PA} & \textbf{CW PA}
 & \textbf{PA} & \textbf{CW PA} \\
\midrule

\multirow{5}{*}{\textbf{Spatial}}
 & Person Initial Position      & 0.64 ± 0.01 & 1.00 ± 0.00 & 0.46 ± 0.01 & 0.73 ± 0.00 & 0.49 ± 0.01 & 0.78 ± 0.01 \\
 & Person Ending Position       & 0.61 ± 0.01 & 1.00 ± 0.00 & 0.43 ± 0.01 & 0.71 ± 0.01 & 0.41 ± 0.01 & 0.67 ± 0.02 \\
 & Goal Initial Position        & 0.80 ± 0.02 & 1.00 ± 0.00 & 0.68 ± 0.03 & 0.85 ± 0.01 & 0.68 ± 0.04 & 0.83 ± 0.05 \\
 & Goal End Position            & 0.77 ± 0.02 & 1.00 ± 0.00 & 0.62 ± 0.02 & 0.82 ± 0.01 & 0.56 ± 0.04 & 0.72 ± 0.05 \\
 & Person End Goal Obstruction  & 0.86 ± 0.01 & 1.00 ± 0.00 & 0.77 ± 0.01 & 0.89 ± 0.00 & 0.80 ± 0.01 & 0.93 ± 0.01 \\
\midrule

\multirow{3}{*}{\textbf{Spatiotemporal}}
 & Robot Moving Direction       & 0.69 ± 0.03 & 1.00 ± 0.00 & 0.52 ± 0.03 & 0.74 ± 0.02 & 0.62 ± 0.04 & 0.87 ± 0.03 \\
 & Person Distance Change       & 0.63 ± 0.01 & 1.00 ± 0.00 & 0.46 ± 0.01 & 0.74 ± 0.00 & 0.48 ± 0.01 & 0.76 ± 0.02 \\
 & Person Goal Obstruction      & 0.83 ± 0.01 & 1.00 ± 0.00 & 0.73 ± 0.01 & 0.88 ± 0.00 & 0.78 ± 0.01 & 0.94 ± 0.01 \\
\midrule

\multirow{7}{*}{\textbf{Social}}
 & Robot Affected by Person          & 0.82 ± 0.01 & 1.00 ± 0.00 & 0.72 ± 0.01 & 0.87 ± 0.00 & 0.76 ± 0.01 & 0.91 ± 0.01 \\
 & Robot Action to Person            & 0.67 ± 0.01 & 1.00 ± 0.00 & 0.50 ± 0.01 & 0.74 ± 0.01 & 0.57 ± 0.01 & 0.82 ± 0.02 \\
 & Person Affected by Robot          & 0.84 ± 0.01 & 1.00 ± 0.00 & 0.74 ± 0.01 & 0.88 ± 0.00 & 0.79 ± 0.01 & 0.94 ± 0.01 \\
 & Person Action to Robot            & 0.72 ± 0.01 & 1.00 ± 0.00 & 0.56 ± 0.01 & 0.77 ± 0.01 & 0.67 ± 0.01 & 0.93 ± 0.01 \\
 & Robot Affected by Person at end   & 0.84 ± 0.01 & 1.00 ± 0.00 & 0.73 ± 0.01 & 0.88 ± 0.00 & 0.79 ± 0.01 & 0.94 ± 0.01 \\
 & Robot Action to Person at end     & 0.70 ± 0.01 & 1.00 ± 0.00 & 0.53 ± 0.01 & 0.75 ± 0.01 & 0.67 ± 0.01 & 0.94 ± 0.01 \\
 & Person Action to Robot at end     & 0.71 ± 0.01 & 1.00 ± 0.00 & 0.54 ± 0.01 & 0.76 ± 0.01 & 0.70 ± 0.01 & 0.98 ± 0.01 \\
\bottomrule
\end{tabular}
\begin{minipage}{\textwidth}
\label{tab:question_performance_baseline}
\end{minipage}
\end{table}

\FloatBarrier

\subsection{Ablation Experiments}
\label{appendix:ablations}
\begin{table}[htbp!]
\centering
\caption{\textbf{Ablation experiment of querying strategies}. The metric used is Probability of Agreement (PA). The baseline row BEV+CoT represents the VLM's performance with both CoT and BEV prompts enabled, while the subsequent rows show the effects of removing either CoT or BEV components.}
\small
\begin{tabular}{l l c c c}
\hline

\multirow{2}{*}{\textbf{Model}} & \multirow{2}{*}{\textbf{Ablation}} & \textbf{Spatial} & \textbf{Spatiotemporal} & \textbf{Social} \\ 
 & &\textbf{Reasoning} &  \textbf{Reasoning}&\textbf{Reasoning} \\ [0.35ex]
\hline
\noalign{\vskip 4pt}
\multirow{3}{*}{GPT-4o} & CoT+BEV & 0.56 ± 0.01 & 0.51 ± 0.01 & 0.47 ± 0.01 \\ [0.35ex]
 & No CoT  & 0.58 ± 0.01 & 0.53 ± 0.01 & 0.35 ± 0.01 \\ [0.35ex]
 & No BEV & 0.51 ± 0.01 & 0.44 ± 0.01 & 0.42 ± 0.01 \\ [0.35ex]
 \hline
 \noalign{\vskip 4pt}
 \multirow{3}{*}{LLaVa-Next-Video} & CoT+BEV & 0.35 ± 0.01 & 0.58 ± 0.01 & 0.48 ± 0.01 \\ [0.35ex]
 & No CoT  & 0.35 ± 0.01 & 0.58 ± 0.01 & 0.38 ± 0.01 \\ [0.35ex]
 & No BEV & 0.35 ± 0.01 & 0.61 ± 0.01 & 0.46 ± 0.01 \\ [0.35ex]
  \hline
  \noalign{\vskip 4pt}
 \multirow{3}{*}{Gemini 2.0} & CoT+BEV & 0.55 ± 0.01 & 0.46 ± 0.01 & 0.63 ± 0.01 \\ [0.35ex]
 & No CoT  & 0.56 ± 0.01 & 0.48 ± 0.01 & 0.58 ± 0.01 \\ [0.35ex]
 & No BEV & 0.56 ± 0.01 & 0.46 ± 0.01 & 0.64 ± 0.01 \\ [0.35ex]
\hline
\end{tabular} 
\label{tab:ablation_1}
\end{table}

To understand the impact of specific querying strategies on model performance, we conducted ablation experiments, systematically removing components such as chain-of-thought (CoT) reasoning and BEV prompts. Table~\ref{tab:ablation_1} summarizes how these ablations affect PA in spatial, spatio-temporal, and social reasoning tasks.

{\bf CoT reasoning.} The results indicate that removing the CoT component does not significantly affect spatial and spatiotemporal reasoning performance. However, the removal of CoT leads to a notable decrease in social reasoning performance across all models. We hypothesize that social reasoning tasks more often require multi-step reasoning to which CoT can help structure complex chains of inference.

{\bf BEV visual prompts.} The results from removing BEV prompts indicate that there is not a significant effect across the capabilities for LLaVa-Next-Video and Gemini 2.0, but provides a notable decrease in performance for GPT-4o across all capabilities. While these results may not indicate a clear winner for all models, it suggests that prompt design remains an open question which needs to be further studied, an endeavor that can be pursued using our benchmark.

{\bf Spatial Reasoning's Affect on Performance.} We ran an additional experiment to see if a lack of strong performance for spatial and spatiotemporal reasoning was affecting performance on social reasoning questions. Table \ref{tab:using_gt_spatial} shows the results of running this experiment where we used the human consensus answer's for the answers for spatial and spatiotemporal reasoning questions for the VLM, which was also provided as chain-of-thought reasoning to the VLM in the form of context; the VLM was then evaluated on social reasoning questions. These results indicate that a strong spatial and spatiotemporal reasoning capabilities can lead to significantly better performance on social reasoning questions.  The ``Person Goal Obstruction" question may provide sufficient information for the VLM to easily answer the ``Robot Affected By Person" question, to which we run an additional experiment and empirically found that, although it was not as drastic, there were performance gains across all questions. These results indicate that hybrid VLM systems that help VLM's with their weaknesses (such as dedicated perception modules) may be more effective rather than entirely relying on the VLM for all questions.

\begin{table}[!t]
\centering
\caption{\textbf{Gemini ablation experiments when using ground truth spatial and spatiotemporal answers for CoT reasoning.} Our results indicate that better spatial reasoning and spatiotemporal reasoning leads to better performance on social reasoning questions.}
\tiny
\begin{tabular}{llcccccccccc}
\toprule
\multirow{2}{*}{\textbf{}} 
 & \multirow{2}{*}{\textbf{Question Name}} 
 & \multicolumn{2}{c}{\textbf{CoT}} 
 & \multicolumn{2}{c}{\textbf{CoT with Ground-Truth Spatial(Temporal) Reasoning}} \\
\cmidrule(lr){3-4}
\cmidrule(lr){5-6}
 & 
 & \textbf{PA} & \textbf{CW PA}
 & \textbf{PA} & \textbf{CW PA} \\
\midrule

\multirow{7}{*}{\textbf{}}
& Robot Affected by Person          & 0.64 ± 0.02 & 0.78 ± 0.02 & 0.78 ± 0.01 & 0.94 ± 0.01 \\
& Robot Action to Person            & 0.51 ± 0.01 & 0.75 ± 0.02 & 0.60 ± 0.01 & 0.88 ± 0.01 \\
& Person Affected by Robot          & 0.74 ± 0.01 & 0.88 ± 0.01 & 0.78 ± 0.01 & 0.94 ± 0.01 \\
& Person Action to Robot            & 0.62 ± 0.01 & 0.86 ± 0.02 & 0.65 ± 0.01 & 0.90 ± 0.01 \\
& Robot Affected by Person at end   & 0.72 ± 0.01 & 0.87 ± 0.01 & 0.78 ± 0.01 & 0.93 ± 0.01 \\
& Robot Action to Person at end     & 0.60 ± 0.01 & 0.85 ± 0.02 & 0.65 ± 0.01 & 0.91 ± 0.01 \\
& Person Action to Robot at end     & 0.62 ± 0.01 & 0.87 ± 0.01 & 0.65 ± 0.01 & 0.91 ± 0.01 \\
\bottomrule
\end{tabular}
\label{tab:using_gt_spatial}
\end{table}

\FloatBarrier

\subsection{Rule-Based Baseline Details}
\label{appendix:rule_baseline}
As mentioned in Section \ref{subsec:experiments_process}, we developed a rule-based baseline which uses a set of hand-crafted rules to determine answers for VQA questions.
Although our simple approach demonstrates better performance than VLMs, it is by no means comprehensive and more complex rules can be devised to further push performance. We briefly summarize the simple rules to determine answers for our Rule-Baed baseline:
\begin{itemize}
    \item Spatial Reasoning Position Questions: Determine deviation in the horizontal direction and use it along with cutoff values to determine whether to answer they are to the left, ahead, or behind.
    \item Goal Obstruction Questions: Draw a line from the robot to the goal and a line from the person's trajectory, if the lines intersect, consider the person obstructing the goal.
    \item Person Distance Change: Look at the initial relative position and end relative positions for the person, determine the appropriate answer based on the distance between the two points.
    \item Robot Moving Direction: Use the horizontal deviation between the the initial relative position of the robot and the end relative position to determine if the robot is turning.
    \item Social Reasoning ``Affected" Questions: If the person is obstructing the goal, then answer that the robot will be affected by them.
    \item Social Reasoning ``Action" Questions: If the person is obstructing the goal, then avoid the person. For person action questions, use the same answer as the robot action questions.
\end{itemize}

\label{sec:appendix}

\end{document}